\documentclass[twoside,11pt]{article}
\usepackage{jmlr2e}

\usepackage{hyperref}

\hypersetup{
    colorlinks,
    citecolor=black,
    filecolor=black,
    linkcolor=black,
    urlcolor=black,
    linktoc=all
}

\usepackage{bookmark}
\usepackage{url}
\usepackage{graphicx}
\usepackage{amsfonts}
\usepackage{amsmath}
\usepackage{booktabs}
\usepackage{listings}
\usepackage{amsmath}
\usepackage{bbm}
\usepackage{xcolor}
\usepackage{enumitem}

\usepackage{booktabs}

\definecolor{codegreen}{rgb}{0,0.5,0}
\definecolor{codeblue}{rgb}{0.25,0.5,0.5}
\definecolor{codegray}{rgb}{0.6,0.6,0.6}

\makeatletter
\newcommand{\printfnsymbol}[1]{%
  \textsuperscript{\@fnsymbol{#1}}%
}
\makeatother

\usepackage{lastpage}
\jmlrheading{23}{2022}{1-\pageref{LastPage}}{8/21; Revised
3/22}{4/22}{21-0998}{William Fedus, Barret Zoph and Noam Shazeer}
\ShortHeadings{Switch Transformers}{Fedus, Zoph and Shazeer}

\firstpageno{1}

\begin{document}

\title{Switch Transformers: Scaling to Trillion Parameter Models with Simple and Efficient Sparsity}


\author{\name William Fedus\thanks{Equal contribution.} \\
\email liamfedus@google.com
\AND    
\name Barret Zoph\printfnsymbol{1} \\
\email barretzoph@google.com
\AND
\name Noam Shazeer \\
\email noam@google.com \\
\addr Google, Mountain View, CA 94043, USA
}

\editor{Alexander Clark}

\maketitle

\begin{abstract}%
In deep learning, models typically reuse the same parameters for all inputs.
Mixture of Experts (MoE) models defy this and instead select \emph{different} parameters for each incoming example.
The result is a sparsely-activated model---with an outrageous number of parameters---but a constant computational cost.
However, despite several notable successes of MoE, widespread adoption has been hindered by  complexity, communication costs, and training instability.
We address these with the introduction of the Switch Transformer.
We simplify the MoE routing algorithm and design intuitive improved models with reduced communication and computational costs.
Our proposed training techniques mitigate the instabilities, and we show large sparse models may be trained, for the first time, with lower precision (bfloat16) formats.
We design models based off T5-Base and T5-Large \citep{raffel2019exploring} to obtain up to 7x increases in pre-training speed with the same computational resources.
These improvements extend into multilingual settings where we measure gains over the mT5-Base version across all 101 languages.
Finally, we advance the current scale of language models by pre-training up to trillion parameter models on the ``Colossal Clean Crawled Corpus", and achieve a 4x speedup over the T5-XXL model.\footnote{JAX code for Switch Transformer and all model checkpoints are available at \url{https://github.com/google-research/t5x}}\footnote{Tensorflow code for Switch Transformer is available at  \url{https://github.com/tensorflow/mesh/blob/master/mesh_tensorflow/transformer/moe.py}}
\end{abstract}


\newpage
\setcounter{tocdepth}{2} 
\tableofcontents

\newpage
\section{Introduction}
Large scale training has been an effective path towards flexible and powerful neural language models \citep{radford2018improving,kaplan2020scaling,brown2020language}.
Simple architectures---backed by a generous computational budget, data set size and parameter count---surpass more complicated algorithms \citep{sutton19}.
An approach followed in \citet{radford2018improving, raffel2019exploring, brown2020language} expands the model size of a densely-activated Transformer \citep{vaswani2017attention}.
While effective, it is also extremely computationally intensive \citep{strubell2019energy}. 
Inspired by the success of model scale, but seeking greater computational efficiency, we instead propose a \emph{sparsely-activated} expert model: the Switch Transformer.
In our case the sparsity comes from activating a \emph{subset} of the neural network weights for each incoming example.

\begin{figure}[ht!]
    \centering
    \includegraphics[width=0.49\columnwidth]{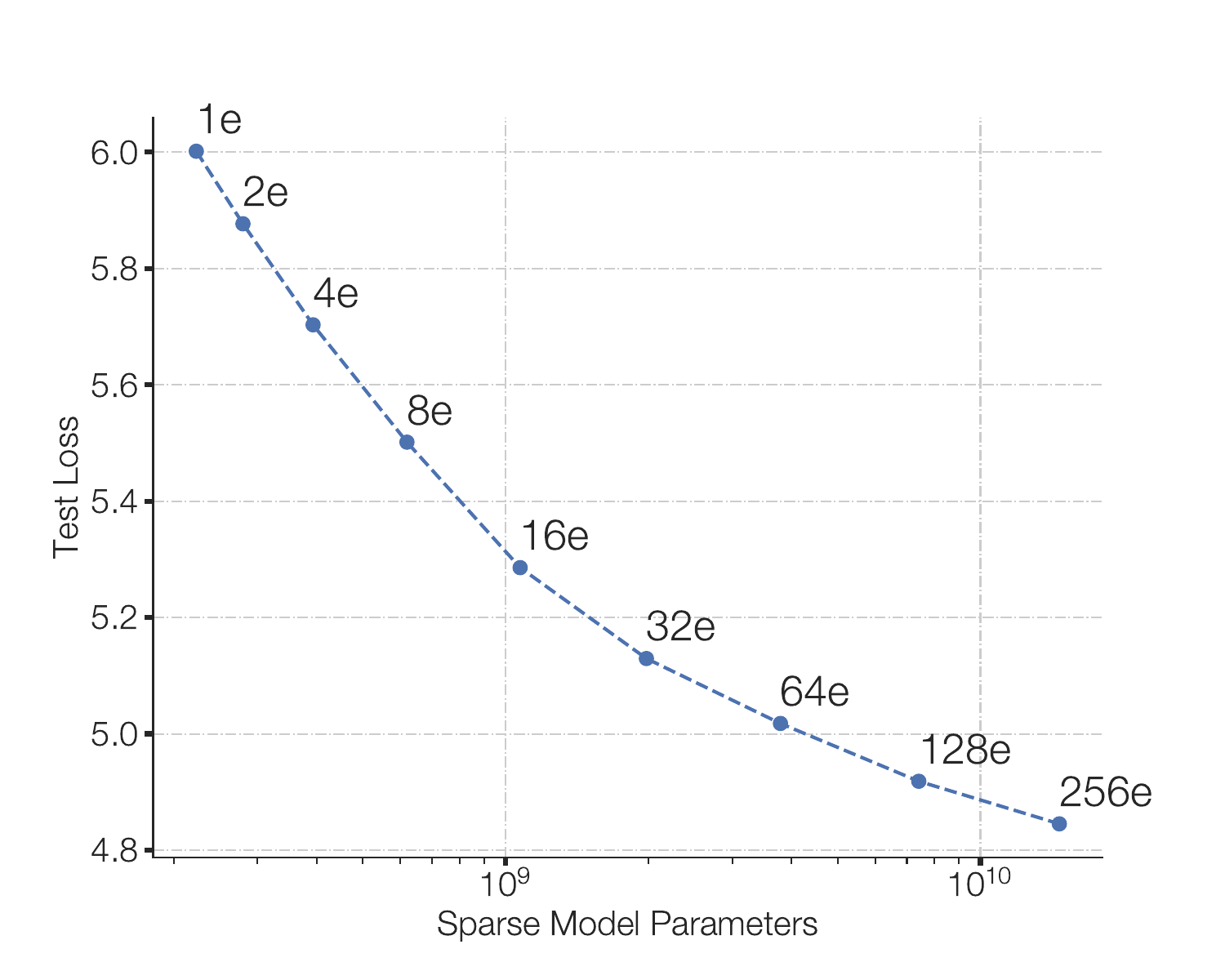}
    \includegraphics[width=0.49\columnwidth]{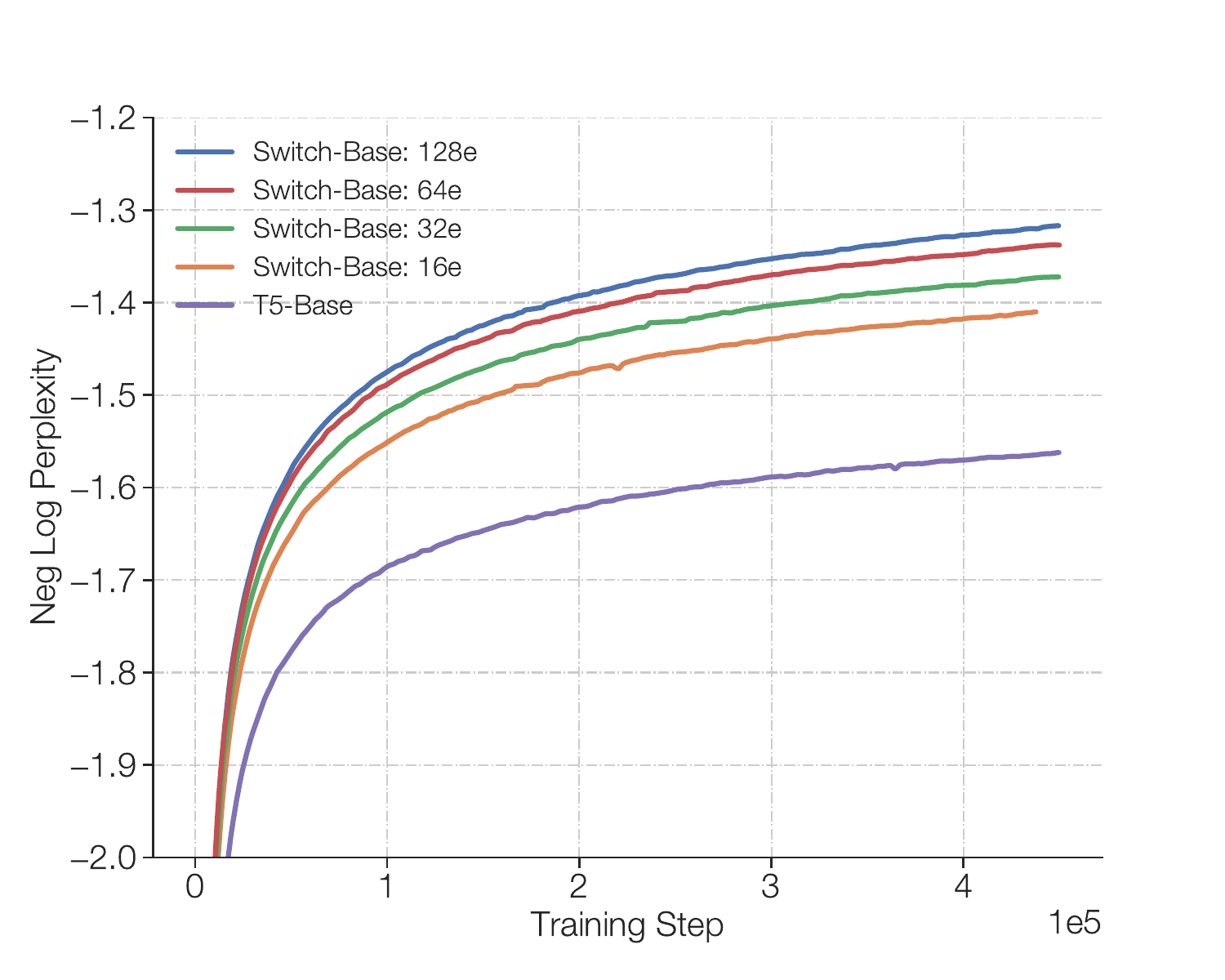}
    \caption{Scaling and sample efficiency of Switch Transformers. Left Plot: Scaling properties for increasingly sparse (more experts) Switch Transformers.  Right Plot: Negative log perplexity comparing Switch Transformers to T5~\citep{raffel2019exploring} models using the same compute budget.}
    \label{fig:summary_figure}
\end{figure}

Sparse training is an active area of research and engineering \citep{gray2017gpu,gale2020sparse}, but as of today, machine learning libraries and hardware accelerators still cater to dense matrix multiplications.
To have an efficient sparse algorithm, we start with the Mixture-of-Expert (MoE) paradigm \citep{jacobs1991adaptive,jordan1994hierarchical,shazeer2017outrageously}, and simplify it to yield training stability and computational benefits.
MoE models have had notable successes in machine translation \citep{shazeer2017outrageously,shazeer2018mesh,lepikhin2020gshard}, however, widespread adoption is hindered by complexity, communication costs, and training instabilities.

We address these issues, and then go beyond translation, to find that these class of algorithms are broadly valuable in natural language.
We measure superior scaling on a diverse set of natural language tasks and across three regimes in NLP: pre-training, fine-tuning and multi-task training.
While this work focuses on scale, we also show that the Switch Transformer architecture not only excels in the domain of supercomputers, but is beneficial even with only a few computational cores.
Further, our large sparse models can be distilled \citep{hinton2015distilling} into small dense versions while preserving 30\% of the sparse model quality gain.
Our contributions are the following:

\begin{itemize}
    \item The Switch Transformer architecture, which simplifies and improves over Mixture of Experts.
    
    \item Scaling properties and a benchmark against the strongly tuned T5 model \citep{raffel2019exploring} where we measure 7x+ pre-training speedups while still using the same FLOPS per token. We further show the improvements hold even with limited computational resources, using as few as two experts.
    
    \item Successful distillation of sparse pre-trained and specialized fine-tuned models into small dense models. We reduce the model size by up to 99\% while preserving 30\% of the quality gains of the large sparse teacher.
    
    \item Improved pre-training and fine-tuning techniques: \textbf{(1)} selective precision training that enables training with lower bfloat16 precision \textbf{(2)} an initialization scheme that allows for scaling to a larger number of experts and \textbf{(3)} increased expert regularization that improves sparse model fine-tuning and multi-task training.
    
    \item A measurement of the pre-training benefits on multilingual data where we find a universal improvement across all 101 languages and with 91\% of languages benefiting from 4x+ speedups over the mT5 baseline \citep{xue2020mt5}.
    
    \item An increase in the scale of neural language models achieved by efficiently combining data, model, and expert-parallelism to create models with up to a trillion parameters.
    These models improve the pre-training speed of a strongly tuned T5-XXL baseline by 4x.
\end{itemize}

\section{Switch Transformer}
The guiding design principle for Switch Transformers is to maximize the parameter count of a Transformer model \citep{vaswani2017attention} in a simple and computationally efficient way. 
The benefit of scale was exhaustively studied in \citet{kaplan2020scaling} which uncovered power-law scaling with model size, data set size and computational budget.
Importantly, this work advocates training large models on relatively small amounts of data as the computationally optimal approach.

Heeding these results, we investigate a fourth axis: increase the \emph{parameter count} while keeping the floating point operations (FLOPs) per example constant.
Our hypothesis is that the parameter count, independent of total computation performed, is a separately important axis on which to scale.
We achieve this by designing a sparsely activated model that efficiently uses hardware designed for dense matrix multiplications such as GPUs and TPUs. 
Our work here focuses on TPU architectures, but these class of models may be similarly trained on GPU clusters.
In our distributed training setup, our sparsely activated layers split \emph{unique} weights on different devices.
Therefore, the weights of the model increase with the number of devices, all while maintaining a manageable memory and computational footprint on each device.
\begin{figure}[t!]
    \centering
    \includegraphics[width=0.9\columnwidth]{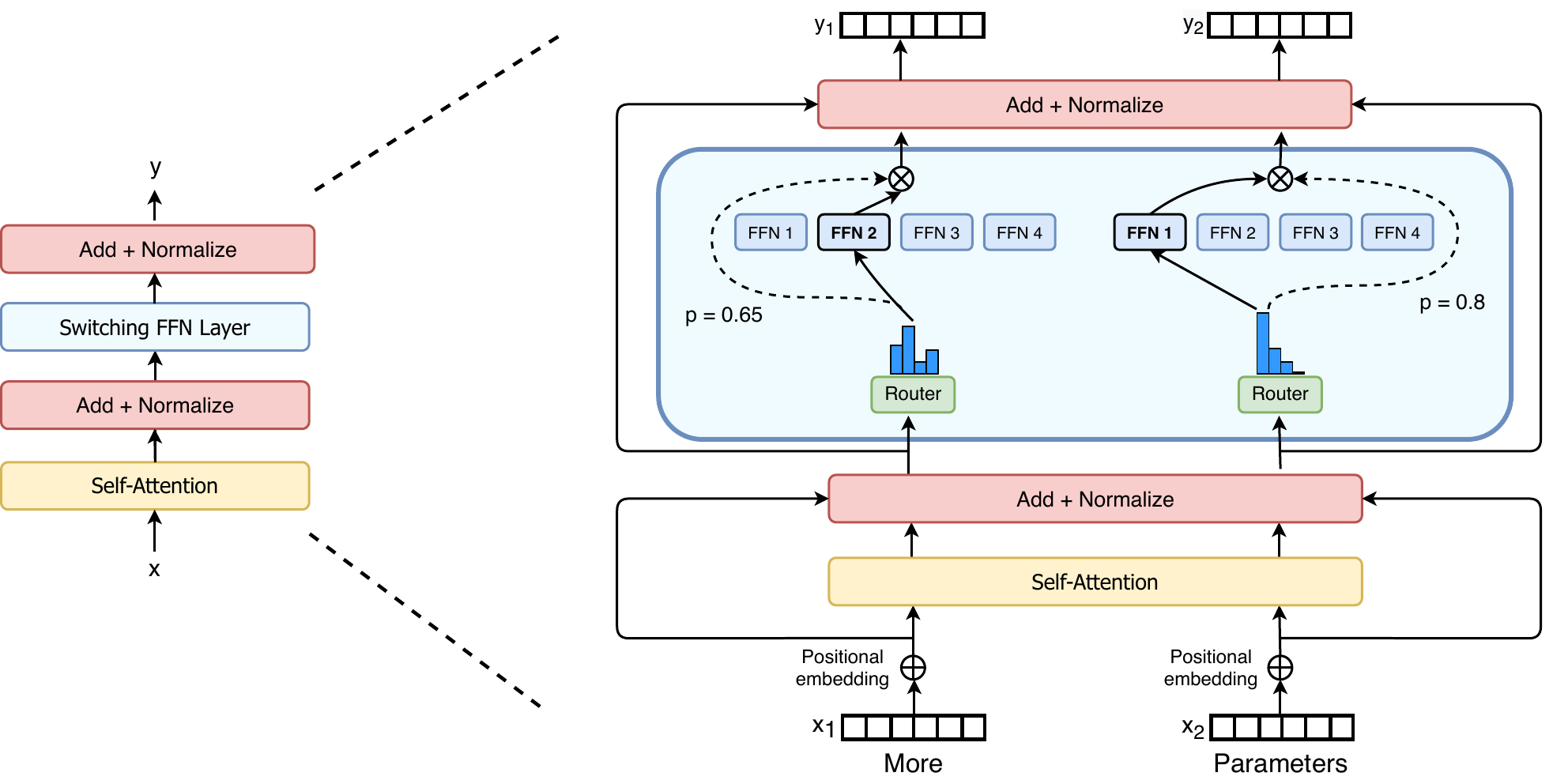}
    \caption{Illustration of a Switch Transformer encoder block. We replace the dense feed forward network (FFN) layer present in the Transformer with a sparse Switch FFN layer (light blue). The layer operates independently on the tokens in the sequence. We diagram two tokens ($x_1=\text{``More"}$ and $x_2=\text{``Parameters"}$ below) being routed (solid lines) across four FFN experts, where the router independently routes each token. The switch FFN layer returns the output of the selected FFN multiplied by the router gate value (dotted-line). }
    \label{fig:experts_attention_performance}
\end{figure}

\subsection{Simplifying Sparse Routing}
\textbf{Mixture of Expert Routing.}
\citet{shazeer2017outrageously} proposed a natural language Mixture-of-Experts (MoE) layer which takes as an input a token representation $x$ and then routes this to the best determined top-$k$ experts, selected from a set $\{E_i(x)\}_{i=1}^N$ of $N$ experts.
The router variable $W_r$ produces logits $h(x) = W_r \cdot x$ which are normalized via a softmax distribution over the available $N$ experts at that layer. The gate-value for expert $i$ is given by,
\begin{equation}
    p_i(x) = \frac{e^{h(x)_i}}{\sum_j^N e^{h(x)_j}}.
\end{equation}
The top-$k$ gate values are selected for routing the token $x$.
If $\mathcal{T}$ is the set of selected top-$k$ indices then the output computation of the layer is the linearly weighted combination of each expert's computation on the token by the gate value,
\begin{equation}\label{eqn: moe_layer}
    y = \sum_{i \in \mathcal{T}} p_i(x) E_i(x). 
\end{equation}

\textbf{Switch Routing: Rethinking Mixture-of-Experts.} 
\citet{shazeer2017outrageously} conjectured that routing to $k>1$ experts was necessary in order to have non-trivial gradients to the routing functions.
The authors intuited that learning to route would not work without the ability to compare at least two experts.
\citet{ramachandran2018diversity} went further to study the top-$k$ decision and found that higher $k$-values in lower layers in the model were important for models with many routing layers.
Contrary to these ideas, we instead use a simplified strategy where we route to only a \emph{single} expert.
We show this simplification preserves model quality, reduces routing computation and performs better.
This $k=1$ routing strategy is later referred to as a Switch layer.
Note that for both MoE and Switch Routing, the gate value $p_i(x)$ in Equation \ref{eqn: moe_layer} permits differentiability of the router.

\begin{figure}[ht!]
    \centering
    \includegraphics[width=1.0\columnwidth]{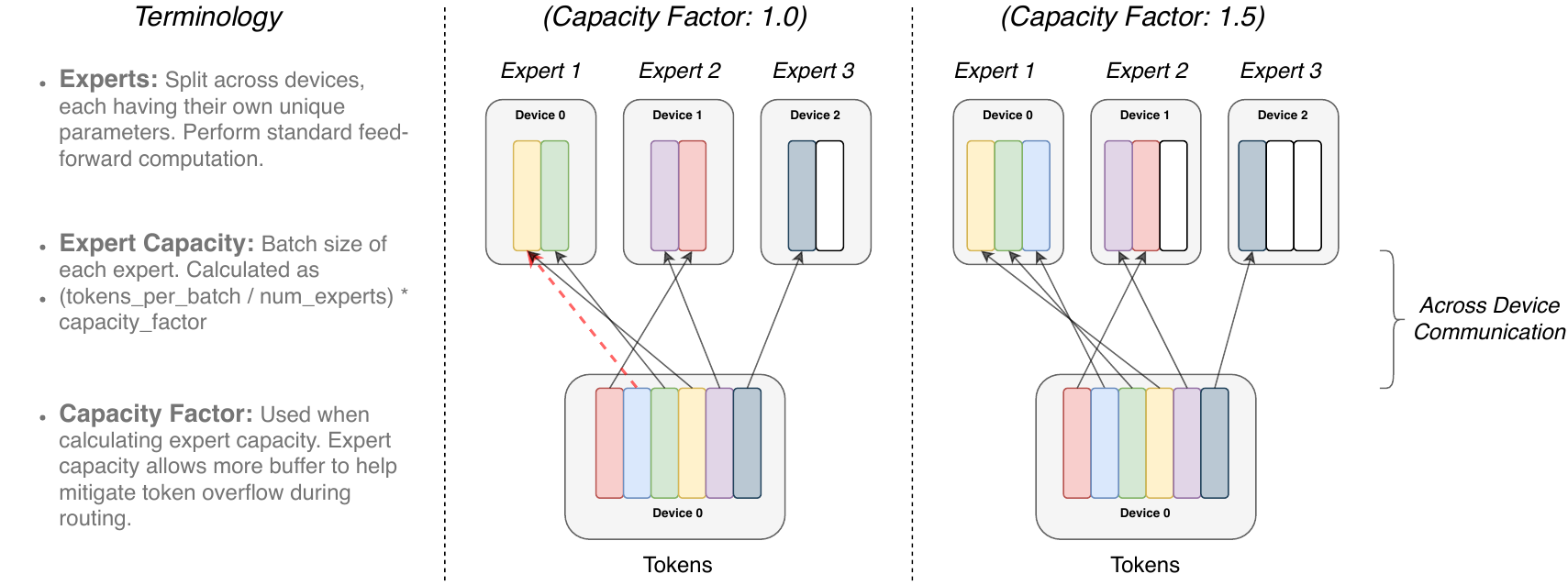}
    \caption{Illustration of token routing dynamics. Each expert processes a fixed batch-size of tokens modulated by the \emph{capacity factor}. Each token is routed to the expert with the highest router probability, but each expert has a fixed batch size of (total\_tokens / num\_experts) $\times$ capacity\_factor. If the tokens are unevenly dispatched then certain experts will overflow (denoted by dotted red lines), resulting in these tokens not being processed by this layer. A larger capacity factor alleviates this overflow issue, but also increases computation and communication costs (depicted by padded white/empty slots).}
    \label{fig:capacity_factor}
\end{figure}

The benefits for the Switch layer are three-fold:
\textbf{(1)} The router computation is reduced as we are only routing a token to a single expert.
\textbf{(2)} The batch size (expert capacity) of each expert can be at least halved since each token is only being routed to a single expert.\footnote{See Section \ref{sec:eff_sparse_rout} for a technical description.}
\textbf{(3)} The routing implementation is simplified and communication costs are reduced.
Figure \ref{fig:capacity_factor} shows an example of routing with different expert capacity factors.
 
\subsection{Efficient Sparse Routing}
\label{sec:eff_sparse_rout}
We use Mesh-Tensorflow (MTF) \citep{shazeer2018mesh} which is a library, with similar semantics and API to Tensorflow \citep{abadi2016tensorflow} that facilitates efficient distributed data and model parallel architectures.
It does so by abstracting the physical set of cores to a logical mesh of processors.
Tensors and computations may then be sharded per named dimensions, facilitating easy partitioning of models across dimensions.
We design our model with TPUs in mind, which require statically declared sizes.
Below we describe our distributed Switch Transformer implementation.

\textbf{Distributed Switch Implementation.}
All of our tensor shapes are statically determined at compilation time, but our computation is \emph{dynamic} due to the routing decisions at training and inference.
Because of this, one important technical consideration is how to set the \emph{expert capacity}.
The expert capacity---the number of tokens each expert computes---is set by evenly dividing the number of tokens in the batch across the number of experts, and then further expanding by a \emph{capacity factor},
\begin{equation}
\text{expert capacity} = \biggr(\frac{\text{tokens per batch}}{\text{number of experts}}\biggr) \times \text{capacity factor}.
\end{equation}
A capacity factor greater than 1.0 creates additional buffer to accommodate for when tokens are not perfectly balanced across experts.
If too many tokens are routed to an expert (referred to later as dropped tokens), computation is skipped and the token representation is passed directly to the next layer through the residual connection.
Increasing the expert capacity is not without drawbacks, however, since high values will result in wasted computation and memory.
This trade-off is explained in Figure \ref{fig:capacity_factor}.
Empirically we find ensuring lower rates of dropped tokens are important for the scaling of sparse expert-models. Throughout our experiments we didn't notice any dependency on the number of experts for the number of tokens dropped (typically $<1\%$). Using the auxiliary load balancing loss (next section) with a high enough coefficient ensured good load balancing.
We study the impact that these design decisions have on model quality and speed  in Table \ref{tab:top1_vs_moe}.


\textbf{A Differentiable Load Balancing Loss.}
To encourage a balanced load across experts we add an auxiliary loss \citep{shazeer2017outrageously,shazeer2018mesh,lepikhin2020gshard}.
As in \cite{shazeer2018mesh,lepikhin2020gshard}, Switch Transformers simplifies the original design in \cite{shazeer2017outrageously} which had separate load-balancing and importance-weighting losses.
For each Switch layer, this auxiliary loss is added to the total model loss during training. 
Given $N$ experts indexed by $i=1$ to $N$ and a batch $\mathcal{B}$ with $T$ tokens, the auxiliary loss is computed as the scaled dot-product between vectors $f$ and $P$,

\begin{equation}  \label{eq:total_loss}
\text{loss} = \alpha \cdot N \cdot \sum_{i=1}^{N} f_i \cdot P_i
\end{equation}

\noindent where $f_i$ is the fraction of tokens dispatched to expert $i$,

\begin{equation} \label{eq:token_sum}
f_i = \frac{1}{T}\sum_{x \in \mathcal{B}} \mathbbm{1} \{\text{argmax}\: p(x) = i\} 
\end{equation}
and $P_i$ is the fraction of the router probability allocated for expert $i$, 
\footnote[2]{A potential source of confusion: $p_i(x)$ is the probability of routing token $x$ to expert $i$. $P_i$ is the probability fraction to expert $i$ across \emph{all tokens} in the batch $\mathcal{B}$.}
\begin{equation} \label{eq:prob_sum}
P_i = \frac{1}{T}\sum_{x \in \mathcal{B}} p_i(x).
\end{equation}
Since we seek uniform routing of the batch of tokens across the $N$ experts, we desire both vectors to have values of $1/N$.
The auxiliary loss of Equation \ref{eq:total_loss} encourages uniform routing since it is minimized under a uniform distribution. The objective can also be differentiated as the $P$-vector is differentiable, but the $f$-vector is not.
The final loss is multiplied by expert count $N$ to keep the loss constant as the number of experts varies since under uniform routing $\sum_{i=1}^N (f_i\cdot P_i) = \sum_{i=1}^N (\frac{1}{N}\cdot \frac{1}{N}) = \frac{1}{N}$.
Finally, a hyper-parameter $\alpha$ is a multiplicative coefficient for these auxiliary losses; throughout this work we use an $\alpha=10^{-2}$ which was sufficiently large to ensure load balancing while small enough to not to overwhelm the primary cross-entropy objective.
We swept hyper-parameter ranges of $\alpha$ from $10^{-1}$ to $10^{-5}$ in powers of 10 and found $10^{-2}$ balanced load quickly without interfering with training loss.

\subsection{Putting It All Together: The Switch Transformer}
Our first test of the Switch Transformer starts with pre-training on the ``Colossal Clean Crawled Corpus'' (C4), introduced in \citep{raffel2019exploring}.
For our pre-training objective, we use a masked language modeling task \citep{taylor1953cloze, fedus2018maskgan, devlin2018bert} where the model is trained to predict missing tokens.
In our pre-training setting, as determined in \citet{raffel2019exploring} to be optimal, we drop out 15\% of tokens and then replace the masked sequence with a single sentinel token.
To compare our models, we record the negative log perplexity.\footnote{We use log base-$e$ for this metric so the units are nats.}
Throughout all tables in the paper, $\uparrow$ indicates that a higher value for that metric is better and vice-versa for $\downarrow$. 
A comparison of all the models studied in this work are in Table \ref{tab: model_params}.

\begin{table}[!th]
\centering
\begin{tabular}{ccccc}
    \toprule
    Model & Capacity & Quality after & Time to Quality & Speed ($\uparrow$) \\ 
      & Factor & 100k steps ($\uparrow$) & Threshold ($\downarrow$) & (examples/sec) \\
      &  & (Neg. Log Perp.) & (hours) &  \\ \hline
    T5-Base & --- & -1.731 & Not achieved$^\dag$ & 1600 \\
    T5-Large & --- & -1.550 & 131.1 & 470 \\ 
    \hline
    MoE-Base & 2.0 & -1.547 & 68.7 & 840 \\ 
    Switch-Base & 2.0 & -1.554 & 72.8 & 860 \\
    \hline
    MoE-Base & 1.25 & -1.559 & 80.7 & 790 \\
    Switch-Base & 1.25 & -1.553 & 65.0 & 910 \\
    \hline
    MoE-Base & 1.0 & -1.572 & 80.1 & 860 \\
    Switch-Base & 1.0 & -1.561 & \textbf{62.8} & 1000 \\ 
    Switch-Base+ & 1.0 & \textbf{-1.534} & 67.6 & 780 \\
    \bottomrule
\end{tabular}
\caption{Benchmarking Switch versus MoE. Head-to-head comparison measuring per step and per time benefits of the Switch Transformer over the MoE Transformer and T5 dense baselines. We measure quality by the negative log perplexity and the time to reach an arbitrary chosen quality threshold of Neg. Log Perp.=-1.50. All MoE and Switch Transformer models use 128 experts, with experts at every other feed-forward layer. For Switch-Base+, we increase the model size until it matches the speed of the MoE model by increasing the model hidden-size from 768 to 896 and the number of heads from 14 to 16. All models are trained with the same amount of computation (32 cores) and on the same hardware (TPUv3). Further note that all our models required pre-training beyond 100k steps to achieve our level threshold of -1.50. $\dag$ T5-Base did not achieve this negative log perplexity in the 100k steps the models were trained.}
\label{tab:top1_vs_moe}
\end{table}

A head-to-head comparison of the Switch Transformer and the MoE Transformer is presented in Table~\ref{tab:top1_vs_moe}.
Our Switch Transformer model is FLOP-matched to `T5-Base' \citep{raffel2019exploring} (same amount of computation per token is applied).
The MoE Transformer, using top-2 routing, has two experts which each apply a separate FFN to each token and thus its FLOPS are larger. 
All models were trained for the same number of steps on identical hardware. Note that the MoE model going from capacity factor 2.0 to 1.25 actually slows down (840 to 790) in the above experiment setup, which is unexpected.\footnote{Note that speed measurements are both a function of the algorithm and the implementation details. Switch Transformer reduces the necessary computation relative to MoE (algorithm), but the final speed differences are impacted by low-level optimizations (implementation).}

We highlight three key findings from Table~\ref{tab:top1_vs_moe}: \textbf{(1)} Switch Transformers outperform both carefully tuned dense models and MoE Transformers on a speed-quality basis.
For a fixed amount of computation and wall-clock time, Switch Transformers achieve the best result. 
\textbf{(2)} The Switch Transformer has a smaller computational footprint than the MoE counterpart. If we increase its size to match the training speed of the MoE Transformer, we find this outperforms all MoE and Dense models on a per step basis as well.
\textbf{(3)} Switch Transformers perform better at lower capacity factors (1.0, 1.25).
Smaller expert capacities are indicative of the scenario in the large model regime where model memory is very scarce and the capacity factor will want to be made as small as possible.

\subsection{Improved Training and Fine-Tuning Techniques}
Sparse expert models may introduce training difficulties over a vanilla Transformer.
Instability can result because of the hard-switching (routing) decisions at each of these layers.
Further, low precision formats like bfloat16 \citep{wang2019bfloat16} can exacerbate issues in the softmax computation for our router.
We describe training difficulties here and the methods we use to overcome them to achieve stable and scalable training.

\textbf{Selective precision with large sparse models.} 
Model instability hinders the ability to train using efficient bfloat16 precision, and as a result, \citet{lepikhin2020gshard} trains with float32 precision throughout their MoE Transformer.
However, we show that by instead \emph{selectively casting} to float32 precision within a localized part of the model, stability may be achieved, without incurring expensive communication cost of float32 tensors. This technique is inline with modern mixed precision training strategies where certain parts of the model and gradient updates are done in higher precision \cite{micikevicius2017mixed}.
Table \ref{tab:selective_precision} shows that our approach permits nearly equal speed to bfloat16 training while conferring the training stability of float32.

\begin{table}[!th]
\centering
    \begin{tabular}{ccc}
    \toprule
        Model  & Quality & Speed  \\
        (precision) & (Neg. Log Perp.) ($\uparrow$) & (Examples/sec) ($\uparrow$) \\ \hline
        Switch-Base (float32) & -1.718  & 1160 \\ 
        Switch-Base (bfloat16) & -3.780 \rlap{[\emph{diverged}]}  & \textbf{1390} \\ 
        Switch-Base (Selective precision) & \textbf{-1.716}  & 1390 \\
    \bottomrule
    \end{tabular}
    \caption{Selective precision. We cast the local routing operations to float32 while preserving bfloat16 precision elsewhere to stabilize our model while achieving nearly equal speed to (unstable) bfloat16-precision training. We measure the quality of a 32 expert model after a fixed step count early in training its speed performance. For both Switch-Base in float32 and with Selective prevision we notice similar learning dynamics.}
    \label{tab:selective_precision}
\end{table}

To achieve this, we cast the router input to float32 precision.
The router function takes the tokens as input and produces the dispatch and combine tensors used for the selection and recombination of expert computation (refer to Code Block \ref{code: router} in the Appendix for details).
Importantly, the float32 precision is only used \emph{within} the body of the router function---on computations local to that device.
Because the resulting dispatch and combine tensors are recast to bfloat16 precision at the end of the function, no expensive float32 tensors are broadcast through all-to-all communication operations, but we still benefit from the increased stability of float32.

\textbf{Smaller parameter initialization for stability}. 
Appropriate initialization is critical to successful training in deep learning and we especially observe this to be true for Switch Transformer.
We initialize our weight matrices by drawing elements from a truncated normal distribution with mean $\mu=0$ and standard deviation $\sigma=\sqrt{s / n}$ where $s$ is a scale hyper-parameter and $n$ is the number of input units in the weight tensor (e.g. fan-in).\footnote{Values greater than two standard deviations from the mean are resampled.}

As an additional remedy to the instability, we recommend reducing the default Transformer initialization scale $s=1.0$ by a factor of 10.
This both improves quality and reduces the likelihood of destabilized training in our experiments.
Table \ref{tab: init} measures the improvement of the model quality and reduction of the variance early in training.
\begin{table}[!th]
\centering
    \begin{tabular}{ccc}
        \toprule
        Model (Initialization scale) & Average Quality & Std. Dev. of Quality \\ 
         & (Neg. Log Perp.) & (Neg. Log Perp.) \\ \hline
        Switch-Base (0.1x-init) & \textbf{-2.72} & \textbf{0.01} \\
        Switch-Base (1.0x-init) & -3.60 & 0.68 \\
        \bottomrule
    \end{tabular}
    \caption{Reduced initialization scale improves stability. Reducing the initialization scale results in better model quality and more stable training of Switch Transformer. Here we record the average and standard deviation of model quality, measured by the negative log perplexity, of a 32 expert model after 3.5k steps (3 random seeds each).}
    \label{tab: init}
\end{table}
We find that the average model quality, as measured by the Neg. Log Perp., is dramatically improved and there is a far reduced variance across runs.
Further, this same initialization scheme is broadly effective for models spanning several orders of magnitude.
We use the same approach to stably train models as small as our 223M parameter baseline to enormous models in excess of one trillion parameters.

\textbf{Regularizing large sparse models.} Our paper considers the common NLP approach of pre-training on a large corpus followed by fine-tuning on smaller downstream tasks such as summarization or question answering.
One issue that naturally arises is overfitting since many fine-tuning tasks have very few examples.
During fine-tuning of standard Transformers, \citet{raffel2019exploring} use dropout \citep{srivastava2014dropout} at each layer to prevent overfitting.
Our Switch Transformers have significantly more parameters than the FLOP matched dense baseline, which can lead to more severe overfitting on these smaller downstream tasks.

\begin{table}[h!]
\begin{center}
\begin{tabular}{cccccccc}
    \toprule
    {Model (dropout)} & {GLUE} & {CNNDM} & {SQuAD} & {SuperGLUE}\\ \hline
    T5-Base (d=0.1) & 82.9 & \textbf{19.6} & 83.5  & 72.4\\
    Switch-Base (d=0.1) & 84.7 & 19.1 & \textbf{83.7} & \textbf{73.0}\\
    Switch-Base (d=0.2) & 84.4 & 19.2 & \textbf{83.9} & \textbf{73.2}\\
    Switch-Base (d=0.3) & 83.9 & 19.6 & 83.4 & 70.7 \\
    Switch-Base (d=0.1, ed=0.4) & \textbf{85.2} & \textbf{19.6} & \textbf{83.7} & \textbf{73.0} \\
    \bottomrule
\end{tabular}
\caption{Fine-tuning regularization results. A sweep of dropout rates while fine-tuning Switch Transformer models pre-trained on 34B tokens of the C4 data set (higher numbers are better). We observe that using a lower standard dropout rate at all non-expert layer, with a much larger dropout rate on the expert feed-forward layers, to perform the best.}
\label{tab:expert_dropout}
\end{center}
\end{table}

We thus propose a simple way to alleviate this issue during fine-tuning: increase the dropout inside the experts, which we name as \emph{expert dropout}.
During fine-tuning we simply increase the dropout rate by a significant amount only at the interim feed-forward computation at each expert layer.
Table \ref{tab:expert_dropout} has the results for our expert dropout protocol.
We observe that simply increasing the dropout across all layers leads to worse performance.
However, setting a smaller dropout rate (0.1) at non-expert layers and a much larger dropout rate (0.4) at expert layers leads to performance improvements on four smaller downstream tasks.

\section{Scaling Properties}\label{sec: scaling}
We present a study of the \emph{scaling properties} of the Switch Transformer architecture during pre-training.
Per \citet{kaplan2020scaling}, we consider a regime where the model is not bottlenecked by either the computational budget or amount of data.
To avoid the data bottleneck, we use the large C4 corpus with over 180B target tokens \citep{raffel2019exploring} and we train until diminishing returns are observed.

The number of experts is the most efficient dimension for scaling our model.
Increasing the experts keeps the computational cost approximately fixed since the model only selects one expert per token, regardless of the number of experts to choose from.
The router must compute a probability distribution over more experts, however, this is a lightweight computation of cost $O(d_{model} \times \text{num experts})$ where $d_{model}$ is the embedding dimension of tokens passed between the layers.
In this section, we consider the scaling properties on a step-basis and a time-basis with a fixed computational budget.

\subsection{Scaling Results on a Step-Basis}
\begin{figure}[ht!]
    \centering
    \includegraphics[width=0.49\columnwidth]{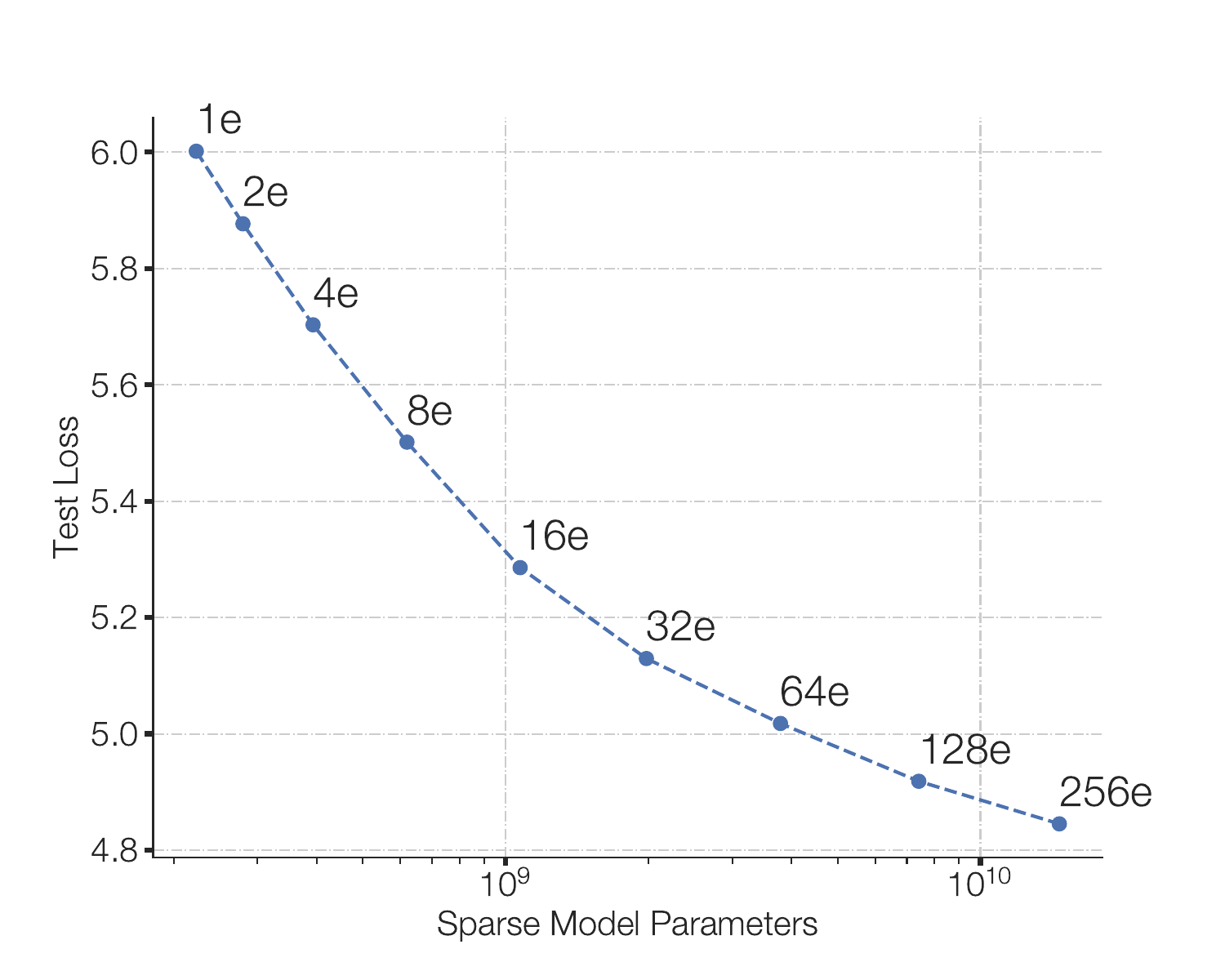}
    \includegraphics[width=0.49\columnwidth]{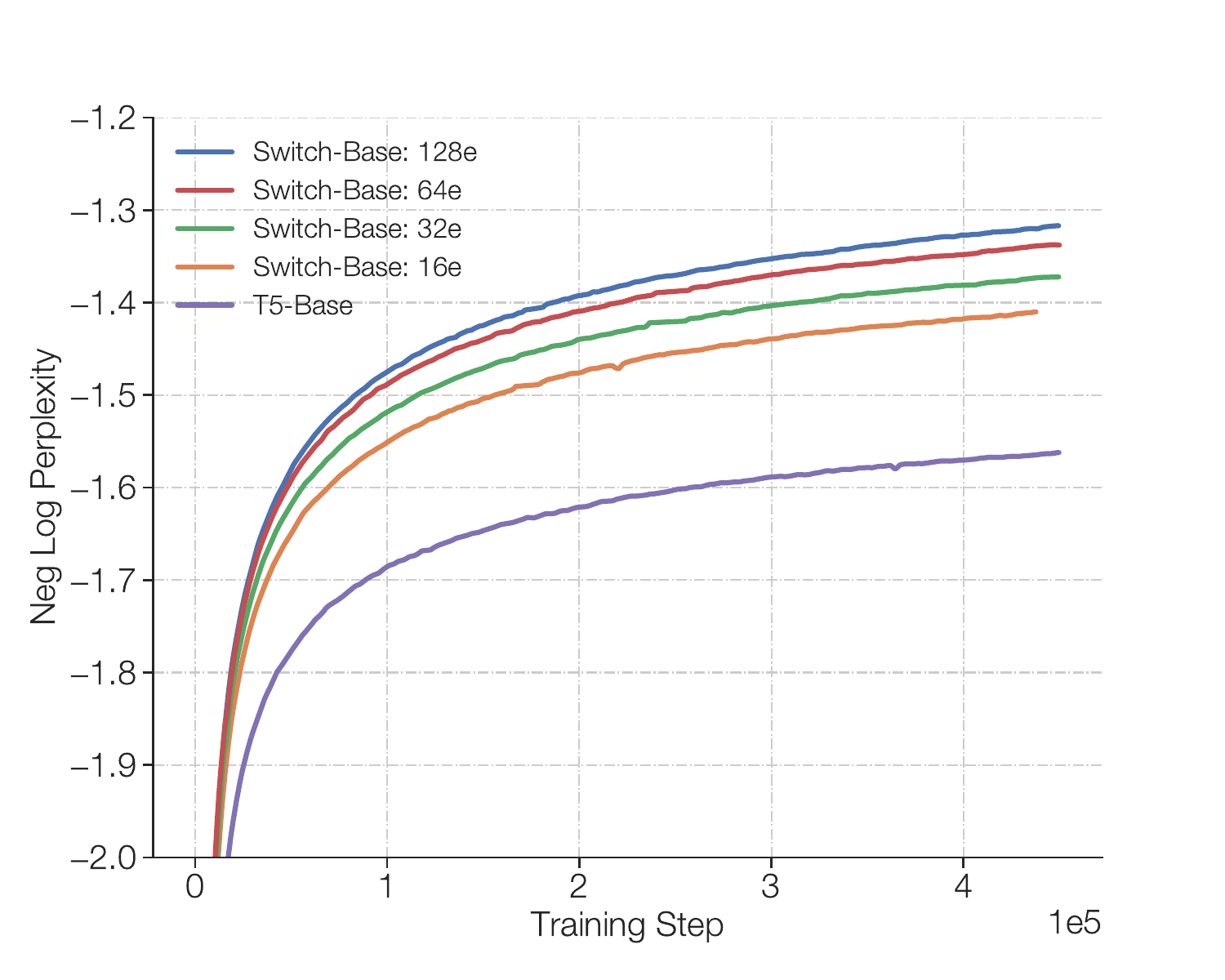}
    \caption{Scaling properties of the Switch Transformer. Left Plot: We measure the quality improvement, as measured by perplexity, as the parameters increase by scaling the number of experts. The top-left point corresponds to the T5-Base model with 223M parameters. Moving from top-left to bottom-right, we double the number of experts from 2, 4, 8 and so on until the bottom-right point of a 256 expert model with 14.7B parameters. Despite all models using an equal computational budget, we observe consistent improvements scaling the number of experts. Right Plot: Negative log perplexity per step sweeping over the number of experts. The dense baseline is shown with the purple line and we note improved sample efficiency of our Switch-Base models.}
    \label{fig:scaling}
\end{figure}

Figure \ref{fig:scaling} demonstrates consistent scaling benefits with the number of experts when training all models for a fixed number of steps.
We observe a clear trend: when keeping the FLOPS per token fixed, having more parameters (experts) speeds up training.
The left Figure demonstrates consistent scaling properties (with fixed FLOPS per token) between sparse model parameters and test loss.
This reveals the advantage of scaling along this additional axis of sparse model parameters.
Our right Figure measures sample efficiency of a dense model variant and four FLOP-matched sparse variants.
We find that increasing the number of experts leads to more sample efficient models.
Our Switch-Base 64 expert model achieves the same performance of the T5-Base model at step 60k at step 450k, which is a ~7.5x speedup in terms of step time.
In addition, consistent with the findings of \citet{kaplan2020scaling}, we find that larger models are also more \emph{sample efficient}---learning more quickly for a fixed number of observed tokens.

\subsection{Scaling Results on a Time-Basis}
Figure \ref{fig:scaling} demonstrates that on a step basis, as we increase the number of experts, the performance consistently improves.
While our models have roughly the same amount of FLOPS per token as the baseline, our Switch Transformers incurs additional communication costs across devices as well as the extra computation of the routing mechanism.
Therefore, the increased sample efficiency observed on a step-basis doesn't necessarily translate to a better model quality as measured by wall-clock.
This raises the question: 

\emph{For a fixed training duration and computational budget, should one train a dense or a sparse model?}

\begin{figure}[!ht]
    \centering
    \includegraphics[width=0.65\columnwidth]{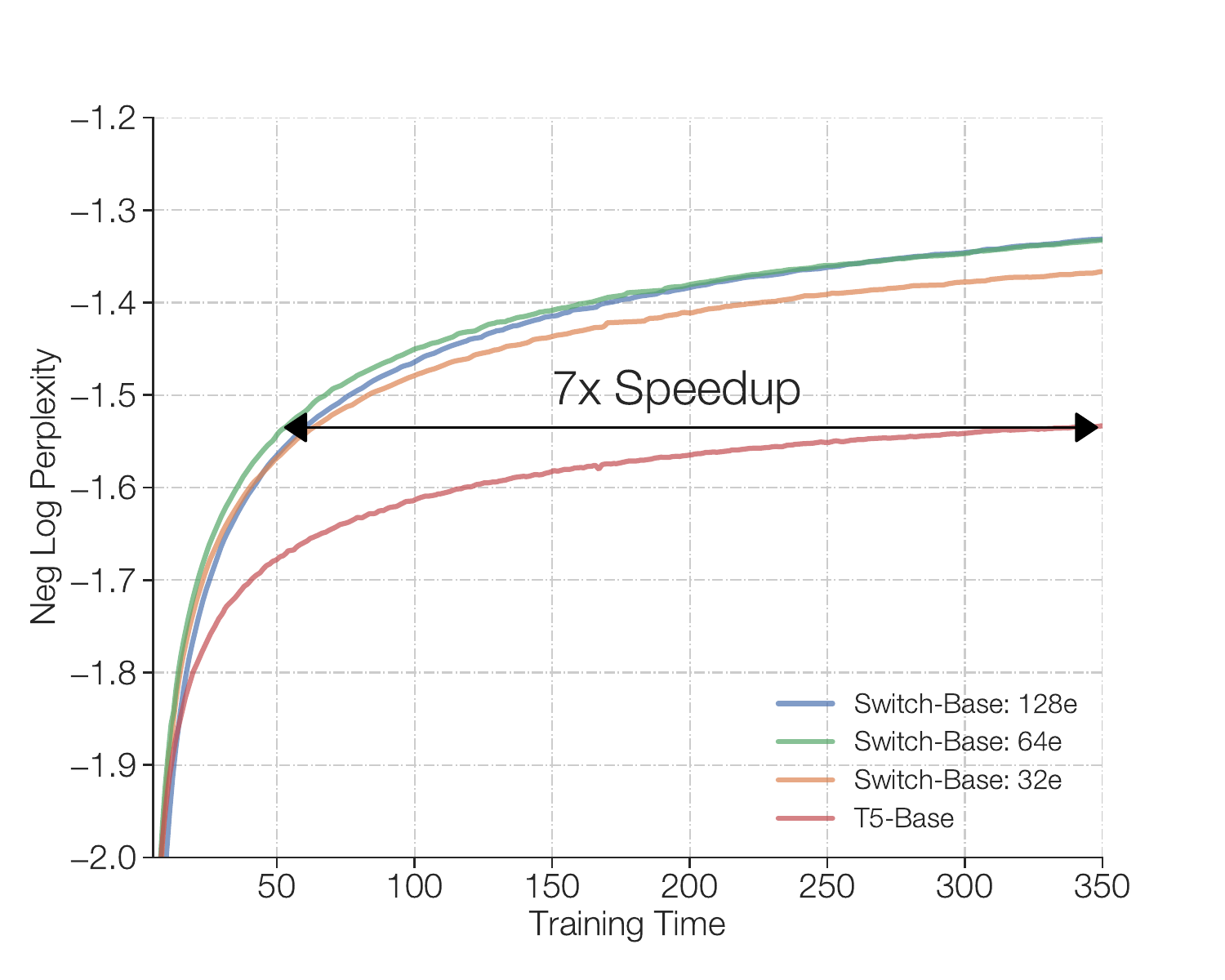} 
    \caption{Speed advantage of Switch Transformer. All models trained on 32 TPUv3 cores with equal FLOPs per example. For a fixed amount of computation and training time, Switch Transformers significantly outperform the dense Transformer baseline. Our 64 expert Switch-Base model achieves the same quality in \emph{one-seventh} the time of the T5-Base and continues to improve.}
    \label{fig:speed_quality_pareto}
\end{figure}

Figures \ref{fig:speed_quality_pareto} and \ref{fig:dense_vs_expert_scaling} address this question.
Figure \ref{fig:speed_quality_pareto} measures the pre-training model quality as a function of time.
For a fixed training duration and computational budget, Switch Transformers yield a substantial speed-up.
In this setting, our Switch-Base 64 expert model trains in \emph{one-seventh} the time that it would take the T5-Base to get similar perplexity.

\subsection{Scaling Versus a Larger Dense Model}
\begin{figure}[ht!]
    \centering
    \includegraphics[width=0.49\columnwidth]{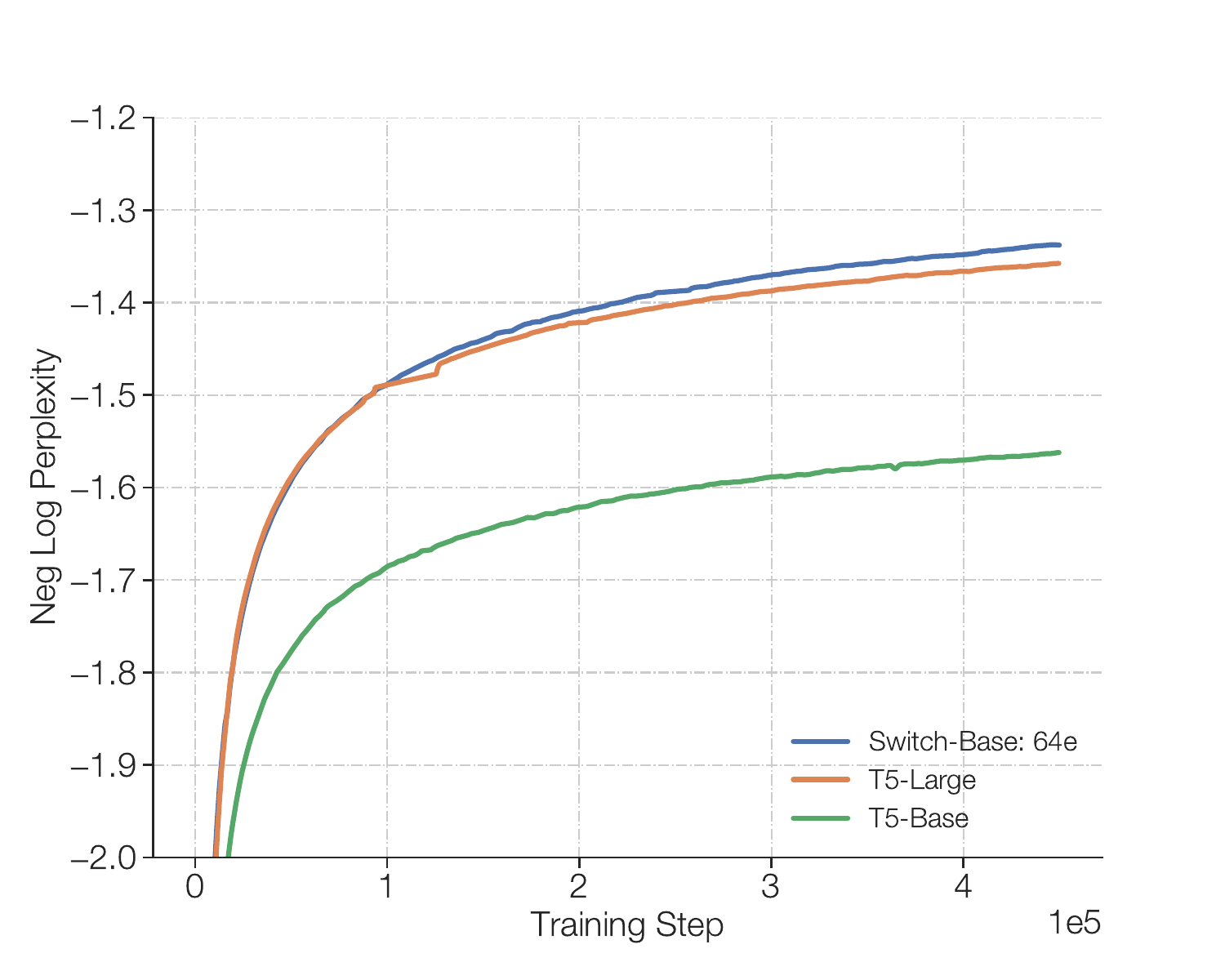}
    \includegraphics[width=0.49\columnwidth]{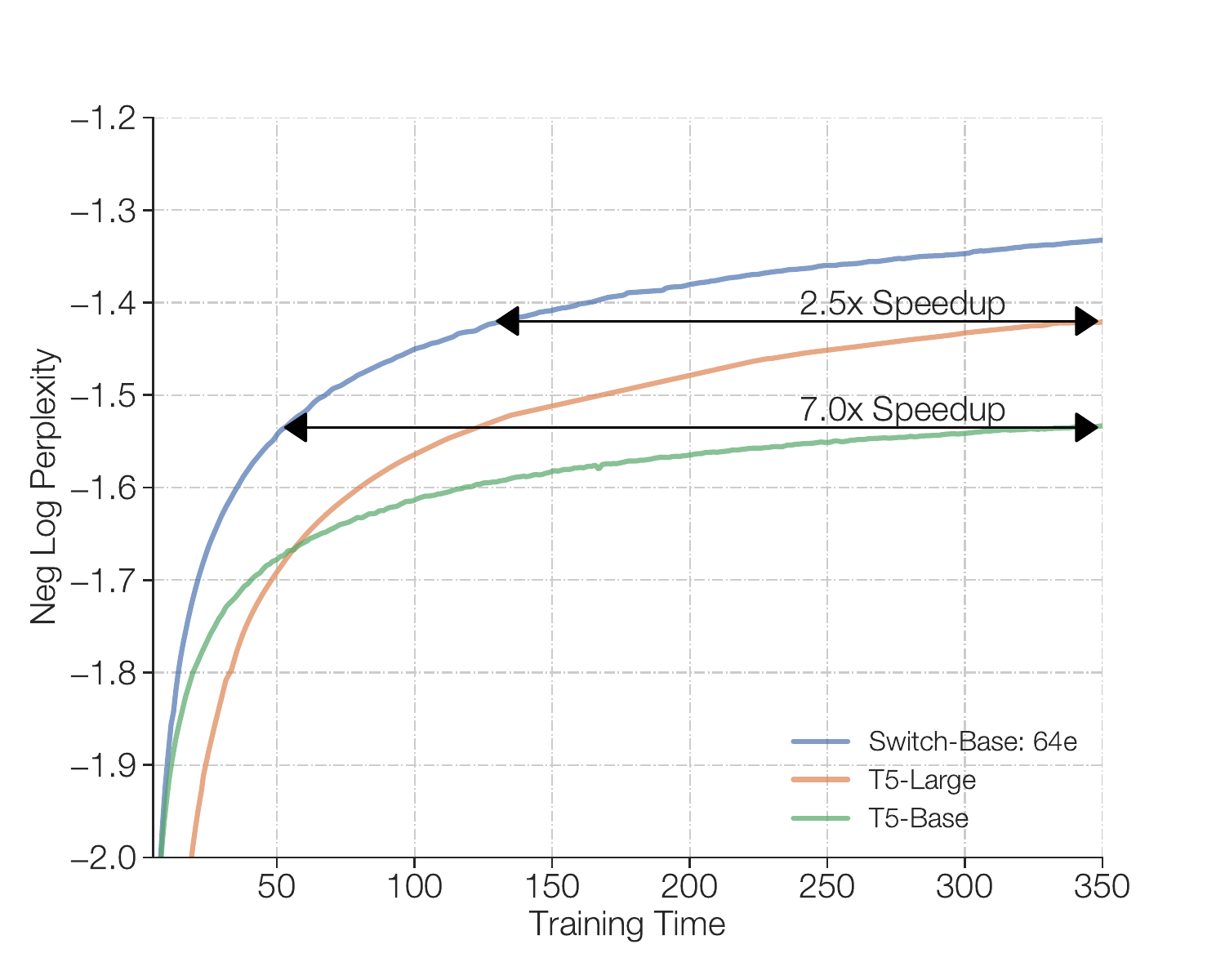}
    \caption{Scaling Transformer models with Switch layers or with standard dense model scaling. Left Plot: Switch-Base is more sample efficient than both the T5-Base, and  T5-Large variant, which applies 3.5x more FLOPS per token. Right Plot: As before, on a wall-clock basis, we find that Switch-Base is still faster, and yields a 2.5x speedup over T5-Large.}
    \label{fig:dense_vs_expert_scaling}
\end{figure}

The above analysis shows that a computationally-matched dense model is outpaced by its Switch counterpart.
Figure \ref{fig:dense_vs_expert_scaling} considers a different scenario: what if we instead had allocated our resources to a larger dense model?
We do so now, measuring Switch-Base against the next strong baseline, \emph{T5-Large}.
But despite T5-Large applying 3.5x more FLOPs per token, Switch-Base is still more sample efficient and yields a 2.5x speedup.
Furthermore, more gains can be had simply by designing a new, larger sparse version, Switch-Large, which is FLOP-matched to T5-Large.
We do this and demonstrate superior scaling and fine-tuning in the following section.

\section{Downstream Results}
Section \ref{sec: scaling} demonstrated the superior scaling properties while pre-training, but we now validate that these gains translate to improved language learning abilities on downstream tasks.
We begin by fine-tuning on a diverse set of NLP tasks.
Next we study reducing the memory footprint of our sparse models by over 90\% by distilling into small---and easily deployed---dense baselines.
Finally, we conclude this section measuring the improvements in a multi-task, multilingual setting, where we show that Switch Transformers are strong multi-task learners, improving over the multilingual T5-base model across all 101 languages.

\subsection{Fine-Tuning}
\textbf{Baseline and Switch models used for fine-tuning.} 
Our baselines are the highly-tuned 223M parameter T5-Base model and the 739M parameter T5-Large model \citep{raffel2019exploring}.
For both versions, we design a FLOP-matched Switch Transformer, with many more parameters, which is summarized in Table \ref{tab: model_params}.\footnote{FLOPS are calculated for the forward pass as done in~\cite{kaplan2020scaling}.}
Our baselines differ slightly from those in \cite{raffel2019exploring} because we pre-train on an improved C4 corpus which removes intra-example text duplication and thus increases the efficacy as a pre-training task \cite{lee2021deduplicating}.
In our protocol we pre-train with $2^{20}$ (1,048,576) tokens per batch for 550k steps amounting to 576B total tokens.
We then fine-tune across a diverse set of tasks using a dropout rate of 0.1 for all layers except the Switch layers, which use a dropout rate of 0.4 (see Table \ref{tab:expert_dropout}).
We fine-tune using a batch-size of 1M for 16k steps and for each task, we evaluate model quality every 200-steps and report the peak performance as computed on the validation set.

\textbf{Fine-tuning tasks and data sets.} We select tasks probing language capabilities including question answering, summarization and knowledge about the world.
The language benchmarks GLUE \citep{wang2018glue} and SuperGLUE \citep{wang2019superglue} are handled as composite mixtures with all the tasks blended in proportion to the amount of tokens present in each.
These benchmarks consist of tasks requiring sentiment analysis (SST-2), word sense disambiguation (WIC), sentence similarty (MRPC, STS-B, QQP), natural language inference (MNLI, QNLI, RTE, CB), question answering (MultiRC, RECORD, BoolQ), coreference resolution (WNLI, WSC) and sentence completion (COPA) and sentence acceptability (CoLA).
The CNNDM \citep{cnn2015moritz} and BBC XSum \citep{narayan2018don} data sets are used to measure the ability to summarize articles.
Question answering is probed with the SQuAD data set \citep{rajpurkar2016squad} and the ARC Reasoning Challenge \citep{clark2018think}.
And as in \cite{roberts2020much}, we evaluate the knowledge of our models by fine-tuning on three closed-book question answering data sets:  Natural Questions \citep{kwiatkowski2019natural}, Web Questions \citep{berant2013semantic} and Trivia QA \citep{joshi2017triviaqa}. 
Closed-book refers to questions posed with no supplemental reference or context material. 
To gauge the model's common sense reasoning we evaluate it on the Winogrande Schema Challenge \citep{sakaguchi2020winogrande}.
And finally, we test our model's natural language inference capabilities on the Adversarial NLI Benchmark \citep{nie2019adversarial}.

\begin{table}[h!]
\begin{center}
\begin{tabular}{ccccc}
    \toprule
    {Model}& {GLUE} & {SQuAD} & {SuperGLUE} & {Winogrande (XL)} \\ \hline
    T5-Base & 84.3 & 85.5 & 75.1 & 66.6   \\
    Switch-Base & \textbf{86.7} & \textbf{87.2} & \textbf{79.5} & \textbf{73.3} \\
    T5-Large & 87.8 & 88.1 & 82.7 & 79.1 \\
    Switch-Large & \textbf{88.5} & \textbf{88.6} & \textbf{84.7} & \textbf{83.0}  \\
    \bottomrule
     & & & & \\
    \toprule
    {Model}& {XSum} & {ANLI (R3)} & {ARC Easy} & {ARC Chal.} \\ \hline
    T5-Base & 18.7 & 51.8 & 56.7 & \textbf{35.5}  \\
    Switch-Base & \textbf{20.3} & \textbf{54.0}  & \textbf{61.3} & 32.8  \\
    T5-Large  & 20.9 & 56.6 & \textbf{68.8} & \textbf{35.5} \\
    Switch-Large  & \textbf{22.3} & \textbf{58.6} & 66.0 & \textbf{35.5} \\
    \bottomrule
     & & & &  \\
    \toprule
    {Model} & {CB Web QA} & {CB Natural QA} & {CB Trivia QA} \\ \hline
    T5-Base & 26.6 & 25.8 & 24.5  \\
    Switch-Base  & \textbf{27.4} & \textbf{26.8} & \textbf{30.7}  \\
    T5-Large  & 27.7 & 27.6 & 29.5 & \\
    Switch-Large  & \textbf{31.3} & \textbf{29.5} & \textbf{36.9}\\
    \bottomrule
\end{tabular}
\caption{Fine-tuning results. Fine-tuning results of T5 baselines and Switch models across a diverse set of natural language tests (validation sets; higher numbers are better). We compare FLOP-matched Switch models to the T5-Base and T5-Large baselines. For most tasks considered, we find significant improvements of the Switch-variants. We observe gains across both model sizes and across both reasoning and knowledge-heavy language tasks.}
\label{tab: fine_tuning}
\end{center}
\end{table}

\textbf{Fine-tuning metrics.} The following evaluation metrics are used throughout the paper:
We report the average scores across all subtasks for GLUE and SuperGLUE.
The Rouge-2 metric is used both the CNNDM and XSum.
In SQuAD and the closed book tasks (Web, Natural, and Trivia Questions) we report the percentage of answers exactly matching the target (refer to \cite{roberts2020much} for further details and deficiency of this measure).
Finally, in ARC Easy, ARC Challenge, ANLI, and Winogrande we report the accuracy of the generated responses.

\textbf{Fine-tuning results.} We observe significant downstream improvements across many natural language tasks.
Notable improvements come from SuperGLUE, where we find FLOP-matched Switch variants improve by 4.4 and 2 percentage points over the T5-Base and T5-Large baselines, respectively as well as large improvements in Winogrande, closed book Trivia QA, and XSum.\footnote{Our T5 and Switch models were pre-trained with $2^{20}$ tokens per batch for 550k steps on a revised C4 data set for fair comparisons.}
In our fine-tuning study, the only tasks where we do not observe gains are on the AI2 Reasoning Challenge (ARC) data sets where the T5-Base outperforms Switch-Base on the challenge data set and T5-Large outperforms Switch-Large on the easy data set.
Taken as a whole, we observe significant improvements spanning both reasoning and knowledge-heavy tasks.
This validates our architecture, not just as one that pre-trains well, but can translate quality improvements to downstream tasks via fine-tuning.

\subsection{Distillation}
Deploying massive neural networks with billions, or trillions, of parameters is inconvenient.
To alleviate this, we study distilling \citep{hinton2015distilling} large sparse models into small dense models. 
Future work could additionally study distilling large models into smaller \emph{sparse} models.

\textbf{Distillation techniques.}
In Table~\ref{tab:distillation_ablation} we study a variety of distillation techniques.
These techniques are built off of ~\cite{sanh2019distilbert}, who study distillation methods for BERT models.
We find that initializing the dense model with the non-expert weights yields a modest improvement. 
This is possible since all models are FLOP matched, so non-expert layers will have the same dimensions.
Since expert layers are usually only added at every or every other FFN layer in a Transformer, this allows for many of the weights to be initialized with trained parameters.
Furthermore, we observe a distillation improvement using a mixture of 0.25 for the teacher probabilities and 0.75 for the ground truth label.
By combining both techniques we preserve $\approx$ 30\% of the quality gains from the larger sparse models with only $\approx1/20^{th}$ of the parameters. 
The quality gain refers to the percent of the quality difference between Switch-Base (Teacher) and T5-Base (Student). 
Therefore, a quality gain of 100\% implies the Student equals the performance of the Teacher.

\begin{center}
\begin{table}[h!]
    \centering
\begin{tabular}{lcr}
    \toprule
    {Technique} & {Parameters} & {Quality ($\uparrow$)} \\
    \hline
    T5-Base & 223M & -1.636  \\
    Switch-Base & 3,800M & -1.444  \\
    \hline
    Distillation & 223M &  \textcolor{blue}{(3\%)} -1.631 \\
    + Init. non-expert weights from teacher & 223M & \textcolor{blue}{(20\%)} -1.598  \\
    + 0.75 mix of hard and soft loss & 223M & \textcolor{blue}{(29\%)} -1.580 \\
    \hline
    Initialization Baseline (no distillation) & & \\
    Init. non-expert weights from teacher & 223M & -1.639 \\ 
    \bottomrule
\end{tabular}
\caption{Distilling Switch Transformers for Language Modeling. 
Initializing T5-Base with the non-expert weights from Switch-Base and using a loss from a mixture of teacher and ground-truth labels obtains the best performance. 
We can distill 30\% of the performance improvement of a large sparse model with 100x more parameters back into a small dense model.
For a final baseline, we find no improvement of T5-Base initialized with the expert weights, but trained normally without distillation.}
\label{tab:distillation_ablation}
\end{table}
\end{center}

\textbf{Achievable compression rates.} 
Using our best distillation technique described in Table~\ref{tab:distillation_ablation}, we distill a wide variety of sparse models into dense models.
We distill Switch-Base versions, sweeping over an increasing number of experts, which corresponds to varying between 1.1B to 14.7B parameters.
Through distillation, we can preserve 37\% of the quality gain of the 1.1B parameter model while compressing 82\%.
At the extreme, where we compress the model 99\%, we are still able to maintain 28\% of the teacher's model quality improvement.
    
\begin{table}[h!]
\centering
\begin{tabular}{cc|ccccc}
    \toprule
    & {Dense} & \multicolumn{5}{c}{Sparse} \\
    \hline
     Parameters & 223M & 1.1B & 2.0B & 3.8B & 7.4B & 14.7B \\ 
     \hline
     Pre-trained Neg. Log Perp. ($\uparrow$) & -1.636 & -1.505 & -1.474 & -1.444 & -1.432 & -1.427 \\
     Distilled Neg. Log Perp. ($\uparrow$) & --- & -1.587 & -1.585 & -1.579 & -1.582 & -1.578 \\ 
     Percent of Teacher Performance & --- & 37\% & 32\% & 30 \% & 27 \% & 28 \% \\
     Compression Percent & --- & 82 \% & 90 \% & 95 \% & 97 \% & 99 \% \\
    \bottomrule
    \label{tab: top1_vs_moe}
\end{tabular}
\caption{Distillation compression rates. We measure the quality when distilling large sparse models into a dense baseline. Our baseline, T5-Base, has a -1.636 Neg. Log Perp. quality. In the right columns, we then distill increasingly large sparse models into this same architecture. Through a combination of weight-initialization and a mixture of hard and soft losses, we can shrink our sparse teachers by 95\%+ while preserving 30\% of the quality gain. 
However, for significantly better and larger pre-trained teachers, we expect larger student models would be necessary to achieve these compression rates.
}
\end{table}

\textbf{Distilling a fine-tuned model.}
We conclude this with a study of distilling a fine-tuned sparse model into a dense model.
Table ~\ref{tab:distillation_superglue} shows results of distilling a 7.4B parameter Switch-Base model, fine-tuned on the SuperGLUE task, into the 223M T5-Base.
Similar to our pre-training results, we find we are able to preserve 30\% of the gains of the sparse model when distilling into a FLOP matched dense variant.
One potential future avenue, not considered here, may examine the specific experts being used for fine-tuning tasks and extracting them to achieve better model compression.

\begin{table}[ht!]
\centering
\begin{tabular}{ccc|r}
    \toprule
    {Model} & {Parameters} & {FLOPS} & {SuperGLUE ($\uparrow$)} \\
    \hline
    T5-Base & 223M & 124B & 74.6 \\
    Switch-Base & 7410M & 124B & 81.3 \\
    Distilled T5-Base & 223M & 124B & \textcolor{blue}{(30\%)} 76.6 \\
    \bottomrule
\end{tabular}
\caption{Distilling a fine-tuned SuperGLUE model. We distill a Switch-Base model fine-tuned on the SuperGLUE tasks into a T5-Base model. We observe that on smaller data sets our large sparse model can be an effective teacher for distillation. We find that we again achieve 30\% of the teacher's performance on a 97\% compressed model.}
\label{tab:distillation_superglue}
\end{table}

\subsection{Multilingual Learning}
In our final set of downstream experiments, we measure the model quality and speed tradeoffs while pre-training on a mixture of 101 different languages.
We build and benchmark off the recent work of mT5 \citep{xue2020mt5}, a multilingual extension to T5.
We pre-train on the multilingual variant of the Common Crawl data set (mC4) spanning 101 languages introduced in mT5, but due to script variants within certain languages, the mixture contains 107 tasks.

\begin{figure}[ht!]
    \centering
    \includegraphics[width=\textwidth]{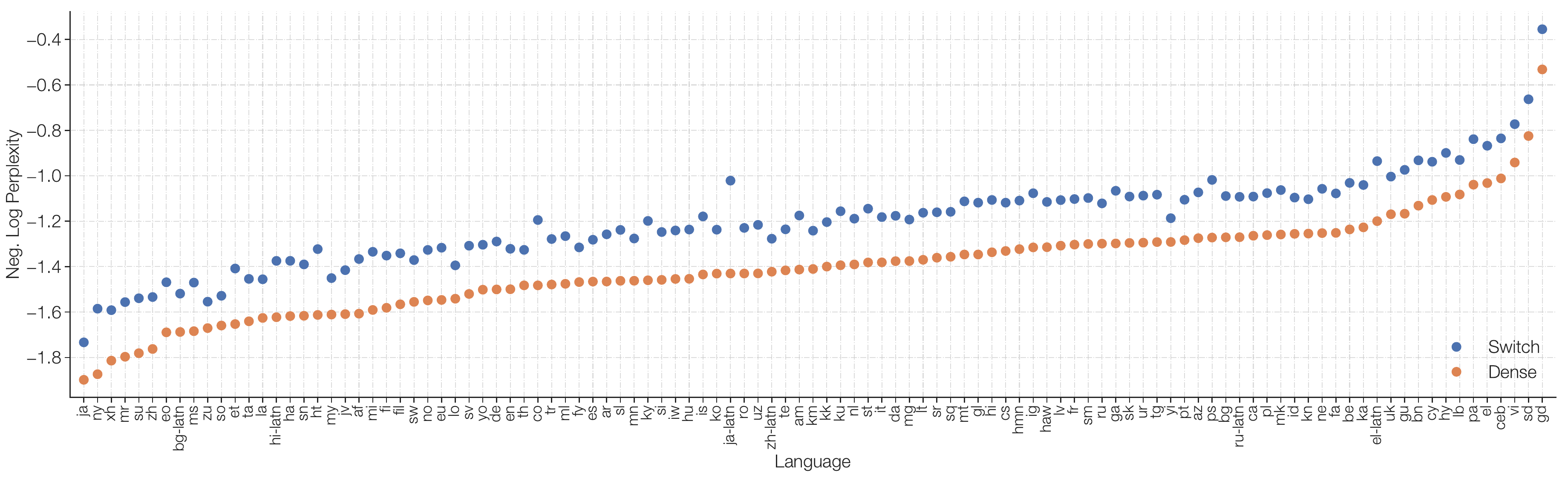}
    \caption{Multilingual pre-training on 101 languages. Improvements of Switch T5 Base model over dense baseline when multi-task training on 101 languages. We observe Switch Transformers to do quite well in the multi-task training setup and yield improvements on all 101 languages.}
    \label{fig:multilingual}
\end{figure}

\begin{figure}[ht!]
    \centering
    \includegraphics[width=0.5\textwidth]{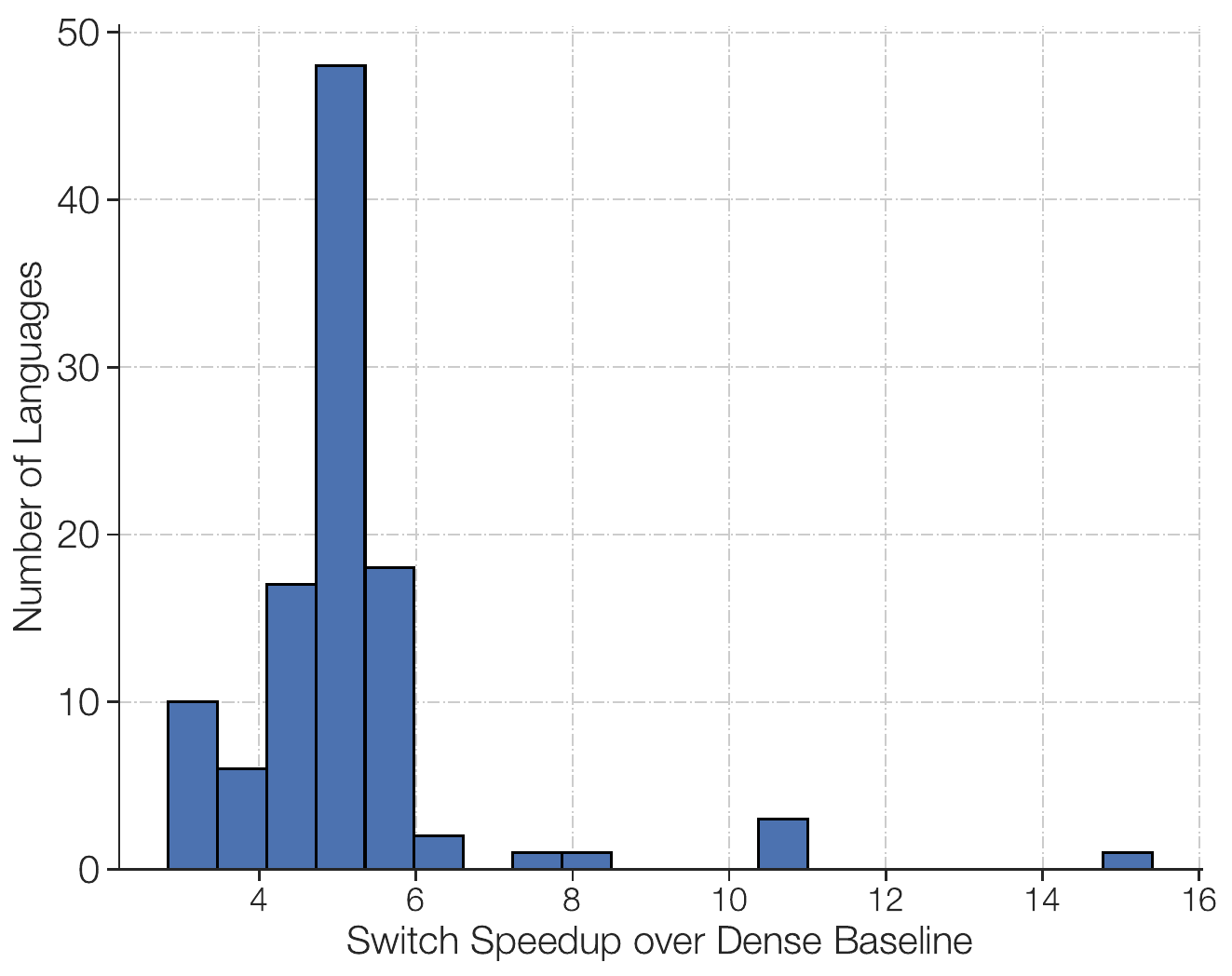}
    \caption{Multilingual pre-training on 101 languages. We histogram for each language, the step speedup of Switch Transformers over the FLOP matched T5 dense baseline to reach the same quality. Over all 101 languages, we achieve a mean step speed-up over mT5-Base of 5x and, for 91\% of languages, we record a 4x, or greater, speedup to reach the final perplexity of mT5-Base.}
    \label{fig:multilingual_speedup}
\end{figure}

In Figure \ref{fig:multilingual} we plot the quality improvement in negative log perplexity for all languages of a FLOP-matched Switch model, mSwitch-Base to the T5 base variant, mT5-Base.
After pre-training both versions for 1M steps, we find that on \emph{all} 101 languages considered, Switch Transformer increases the final negative log perplexity over the baseline.
In Figure \ref{fig:multilingual_speedup}, we present a different view and now histogram the per step \emph{speed-up} of using Switch Transformer over the mT5-Base.\footnote{The speedup on a step basis is computed as the ratio of the number of steps for the baseline divided by the number of steps required by our model to reach that same quality.}
We find a mean speed-up over mT5-Base of 5x and that 91\% of languages achieve at least a 4x speedup.
This presents evidence that Switch Transformers are effective multi-task and multi-lingual learners.

\section{Designing Models with Data, Model, and Expert-Parallelism}
Arbitrarily increasing the number of experts is subject to diminishing returns (Figure \ref{fig:scaling}).
Here we describe \emph{complementary} scaling strategies.
The common way to scale a Transformer is to increase dimensions in tandem, like $d_{model}$ or $d_{ff}$.
This increases both the parameters and computation performed and is ultimately limited by the memory per accelerator.
Once it exceeds the size of the accelerator's memory, single program multiple data (SPMD) model-parallelism can be employed.
This section studies the trade-offs of combining data, model, and expert-parallelism.

\textbf{Reviewing the Feed-Forward Network (FFN) Layer.} 
We use the FFN layer as an example of how data, model and expert-parallelism works in Mesh TensorFlow \citep{shazeer2018mesh} and review it briefly here.
We assume $B$ tokens in the batch, each of dimension $d_{model}$.
Both the input ($x$) and output ($y$) of the FFN are of size [$B$, $d_{model}$] and the intermediate ($h$) is of size [$B$, $d_{ff}$] where $d_{ff}$ is typically several times larger than $d_{model}$.
In the FFN, the intermediate is $h=x W_{in}$ and then the output of the layer is $y = ReLU(h) W_{out}$.
Thus $W_{in}$ and $W_{out}$ are applied independently to each token and have sizes [$d_{model}$, $d_{ff}$] and [$d_{ff}$, $d_{model}$].

\begin{figure}[t!]
    \centering
    \includegraphics[width=1.0\columnwidth]{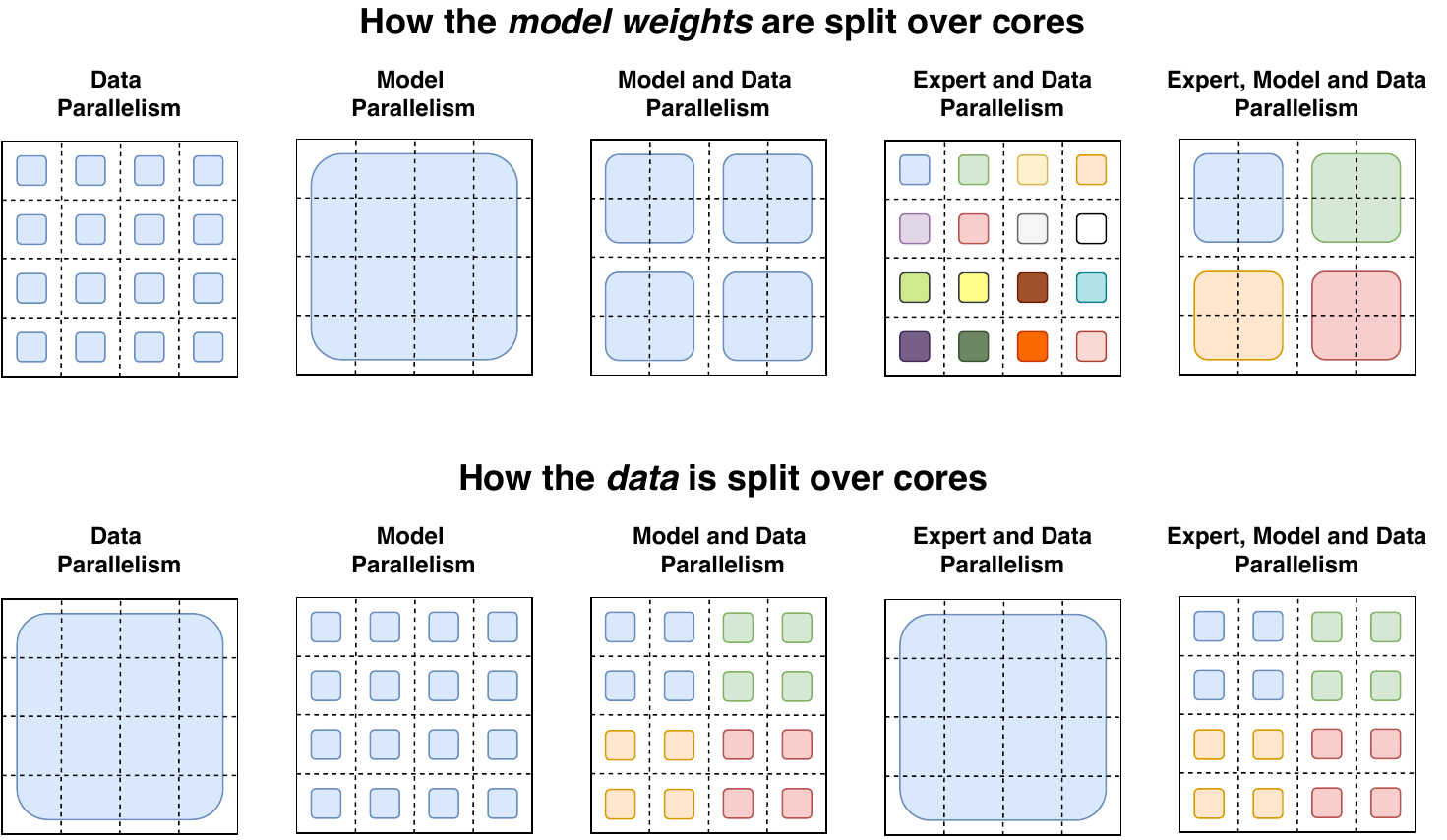}
    \caption{Data and weight partitioning strategies. Each 4$\times$4 dotted-line grid represents 16 cores and the shaded squares are the data contained on that core (either model weights or batch of tokens). We illustrate both how the model weights and the data tensors are split for each strategy. \textbf{First Row:} illustration of how \emph{model weights} are split across the cores. Shapes of different sizes in this row represent larger weight matrices in the Feed Forward Network (FFN) layers (e.g larger $d_{ff}$ sizes). Each color of the shaded squares identifies a unique weight matrix. The number of parameters \emph{per core} is fixed, but larger weight matrices will apply more computation to each token. \textbf{Second Row:} illustration of how the \emph{data batch} is split across cores. Each core holds the same number of tokens which maintains a fixed memory usage across all strategies. The partitioning strategies have different properties of allowing each core to either have the same tokens or different tokens across  cores, which is what the different colors symbolize.}
    \label{fig:core_layouts}
\end{figure}

We describe two aspects of partitioning: how the \emph{weights} and \emph{batches of data} divide over cores, depicted in Figure~\ref{fig:core_layouts}.
We denote all cores available as $N$ which Mesh Tensorflow may then remap into a logical multidimensional mesh of processors.
Here we create a two-dimensional logical mesh, with one dimension representing the number of ways for data-parallel sharding ($n$) and the other, the model-parallel sharding ($m$).
The total cores must equal the ways to shard across both data and model-parallelism, e.g. $N = n \times m$.
To shard the layer across cores, the tensors containing that batch of $B$ tokens are sharded across $n$ data-parallel cores, so each core contains $B / n$ tokens.
Tensors and variables with $d_{ff}$ are then sharded across $m$ model-parallel cores.
For the variants with experts-layers, we consider $E$ experts, each of which can process up to $C$ tokens.

\begin{table}[ht!]
\centering
\begin{tabular}{cl}
    \toprule
    {Term} & {Description} \\
    \hline
    $B$ & Number of tokens in the batch. \\
    $N$ & Number of total cores. \\
    $n$ & Number of ways for data-parallelism sharding. \\
    $m$ & Number of ways for model-parallelism sharding.  \\
    $E$ & Number of experts in Switch layers. \\
    $C$ & Expert capacity, the batch size of each expert.
\end{tabular}
\end{table}

\subsection{Data Parallelism}
When training data parallel models, which is the standard for distributed training, then all cores are allocated to the data-parallel dimension or $n = N, m = 1$.
This has the advantage that no communication is needed until the entire forward and backward pass is finished and the gradients need to be then aggregated across all cores. 
This corresponds to the left-most column of Figure \ref{fig:core_layouts}.

\subsection{Model Parallelism}
We now consider a scenario where all cores are allocated exclusively to the model-parallel dimension and so $n = 1, m = N$.
Now all cores must keep the full $B$ tokens and each core will contain a unique slice of the weights.
For each forward and backward pass, a communication cost is now incurred.
Each core sends a tensor of [$B$, $d_{model}$] to compute the second matrix multiplication $ReLU(h) W_{out}$ because the $d_{ff}$ dimension is partitioned and must be summed over.
As a general rule, whenever a dimension that is partitioned across cores must be summed, then an all-reduce operation is added for both the forward and backward pass.
This contrasts with pure data parallelism where an all-reduce only occurs at the end of the entire forward and backward pass.

\subsection{Model and Data Parallelism}
It is common to mix both model and data parallelism for large scale models, which was done in the largest T5 models ~\citep{raffel2019exploring,xue2020mt5} and in GPT-3 ~\citep{brown2020language}.
With a total of $N = n \times m$ cores, now each core will be responsible for $B / n$ tokens and $d_{ff} / m$ of both the weights and intermediate activation.
In the forward and backward pass each core communicates a tensor of size $[B / n, d_{model}]$ in an all-reduce operation.

\subsection{Expert and Data Parallelism} \label{sec: expert_data_parallelism}
Next we describe the partitioning strategy for expert and data parallelism.
Switch Transformers will allocate all of their cores to the data partitioning dimension $n$, which will also correspond to the number of experts in the model.
For each token per core a router locally computes assignments to the experts.
The output is a binary matrix of size [$n$, $B / n$, $E$, $C$] which is partitioned across the first dimension and determines expert assignment.
This binary matrix is then used to do a gather via matrix multiplication with the input tensor of [$n$, $B / n$, $d_{model}$].
\begin{equation}
    \text{einsum}([n, B / n, d_{model}], [n, B / n, E, C], \text{dimension}=[B / n])
\end{equation}
resulting in the final tensor of shape [$n$, $E$, $C$, $d_{model}$], which is sharded across the first dimension.
Because each core has its own expert, we do an all-to-all communication of size [$E$, $C$, $d_{model}$] to now shard the $E$ dimension instead of the $n$-dimension.
There are additional communication costs of bfloat16 tensors of size $E \times C \times d_{model}$ in the forward pass to analogusly receive the tokens from each expert located on different cores.
See Appendix~\ref{sec:pseudo_code} for a detailed analysis of the expert partitioning code.

\subsection{Expert, Model and Data Parallelism}
\label{sec: expert_model_data_par}
In the design of our best model, we seek to balance the FLOPS per token and the parameter count.
When we scale the number of experts, we increase the number of parameters, but do not change the FLOPs per token.
In order to increase FLOPs, we must also increase the $d_{ff}$ dimension (which also increases parameters, but at a slower rate).
This presents a trade-off: as we increase $d_{ff}$ we will run out of memory per core, which then necessitates increasing $m$. 
But since we have a fixed number of cores $N$, and $N=n \times m$, we must decrease $n$, which forces use of a smaller batch-size (in order to hold tokens per core constant).

When combining both model and expert-parallelism, we will have all-to-all communication costs from routing the tokens to the correct experts along with the internal all-reduce communications from the model parallelism. 
Balancing the FLOPS, communication costs and memory per core becomes quite complex when combining all three methods where the best mapping is empirically determined.
See our further analysis in section~\ref{sec:trillion_results} for how the number of experts effects the downstream performance as well.

\subsection{Towards Trillion Parameter Models}
\label{sec:trillion_results}
Combining expert, model and data parallelism, we design two large Switch Transformer models, one with 395 billion and 1.6 trillion parameters, respectively. 
We study how these models perform on both up-stream pre-training as language models and their downstream fine-tuning performance.
The parameters, FLOPs per sequence and hyper-parameters of the two different models are listed below in Table \ref{tab: model_params}.
Standard hyper-parameters of the Transformer, including $d_{model}$, $d_{ff}$, $d_{kv}$, number of heads and number of layers are described, as well as a less common feature, $FFN_{GEGLU}$, which refers to a variation of the FFN layer where the expansion matrix is substituted with two sets of weights which are non-linearly combined \citep{shazeer2020glu}.

\begin{table}[h!]
\begin{center}
\resizebox{\columnwidth}{!}{
\begin{tabular}{c|ccccccc}
    \toprule
    {Model} & {Parameters} & {FLOPs/seq} & {$d_{model}$} & {$FFN_{GEGLU}$} & {$d_{ff}$} &  {$d_{kv}$} & {Num. Heads} \\ \hline
    T5-Base & 0.2B & 124B & 768  & \checkmark & 2048 & 64 & 12 \\
    T5-Large & 0.7B & 425B & 1024  & \checkmark & 2816 & 64 & 16 \\
    T5-XXL & 11B  & 6.3T & 4096  & \checkmark & 10240 & 64 & 64\\
    \midrule
    Switch-Base & 7B & 124B & 768 & \checkmark & 2048 & 64 & 12 \\
    Switch-Large & 26B & 425B & 1024 & \checkmark & 2816 & 64 & 16 \\
    Switch-XXL & 395B  & 6.3T & 4096 & \checkmark & 10240 & 64 & 64\\
    Switch-C   & 1571B & 890B & 2080  &  & 6144 & 64 & 32 \\
    \bottomrule
    \\
    \toprule
    {Model} & {Expert Freq.} & {Num. Layers} & {Num Experts} & Neg. Log Perp. @250k & Neg. Log Perp. @ 500k\\ \hline
    T5-Base    & -- & 12 & -- & -1.599 & -1.556 \\
    T5-Large     & -- & 24 & -- & -1.402 & -1.350 \\
    T5-XXL     & -- & 24 & --   & -1.147 & -1.095\\
    \midrule
    Switch-Base & $1/2$ & 12 & 128 & -1.370 & -1.306 \\
    Switch-Large & $1/2$ & 24 & 128 & -1.248 & -1.177  \\
    Switch-XXL & $1/2$ & 24 & 64   & \textbf{-1.086} & \textbf{-1.008} \\
    Switch-C   & 1 & 15 & 2048 & -1.096 & -1.043 \\
    \bottomrule
\end{tabular}
}
\caption{Switch model design and pre-training performance. We compare the hyper-parameters and pre-training performance of the T5 models to our Switch Transformer variants. The last two columns record the pre-training model quality on the C4 data set after 250k and 500k steps, respectively. We observe that the Switch-C Transformer variant is 4x faster to a fixed perplexity (with the same compute budget) than the T5-XXL model, with the gap increasing as training progresses.}
\label{tab: model_params}
\end{center}
\end{table}

The Switch-C model is designed using only expert-parallelism, and no model-parallelism, as described earlier in Section \ref{sec: expert_data_parallelism}.
As a result, the hyper-parameters controlling the width, depth, number of heads, and so on, are all much smaller than the T5-XXL model.
In contrast, the Switch-XXL is FLOP-matched to the T5-XXL model, which allows for larger dimensions of the hyper-parameters, but at the expense of additional communication costs induced by model-parallelism (see Section~\ref{sec: expert_model_data_par} for more details).

\textbf{Sample efficiency versus T5-XXL.} In the final two columns of Table \ref{tab: model_params} we record the negative log perplexity on the C4 corpus after 250k and 500k steps, respectively.
After 250k steps, we find both Switch Transformer variants to improve over the T5-XXL version's negative log perplexity by over 0.061.\footnote{This reported quality difference is a lower bound, and may actually be larger. The T5-XXL was pre-trained on an easier C4 data set which included duplicated, and thus easily copied, snippets within examples.}
To contextualize the significance of a gap of 0.061, we note that the T5-XXL model had to train for an \emph{additional} 250k steps to increase 0.052.
The gap continues to increase with additional training, with the Switch-XXL model out-performing the T5-XXL by 0.087 by 500k steps.

\textbf{Training instability.} However, as described in the introduction, large sparse models can be unstable, and as we increase the scale, we encounter some sporadic issues.
We find that the larger Switch-C model, with 1.6T parameters and 2048 experts, exhibits no training instability at all.
Instead, the Switch XXL version, with nearly 10x larger FLOPs per sequence, is sometimes unstable.
As a result, though this is our better model on a step-basis, we do not pre-train for a full 1M steps, in-line with the final reported results of T5 \citep{raffel2019exploring}.

\textbf{Reasoning fine-tuning performance.} As a preliminary assessment of the model quality, we use a Switch-XXL model partially pre-trained on 503B tokens, or approximately half the text used by the T5-XXL model.
Using this checkpoint, we conduct multi-task training for efficiency, where all tasks are learned jointly, rather than individually fine-tuned.
We find that SQuAD accuracy on the validation set increases to 89.7 versus state-of-the-art of 91.3.
Next, the average SuperGLUE test score is recorded at 87.5 versus the T5 version obtaining a score of 89.3 compared to the state-of-the-art of 90.0 \citep{wang2019superglue}.
On ANLI \citep{nie2019adversarial}, Switch XXL improves over the prior state-of-the-art to get a 65.7 accuracy versus the prior best of 49.4 \citep{yang2020xlnet}. 
We note that while the Switch-XXL has state-of-the-art Neg. Log Perp. on the upstream pre-training task, its gains have not yet fully translated to SOTA downstream performance.
We study this issue more in Appendix~\ref{app: upstream_downstream}.

\textbf{Knowledge-based fine-tuning performance.} Finally, we also conduct an early examination of the model's knowledge with three closed-book knowledge-based tasks: Natural Questions, WebQuestions and TriviaQA, without additional pre-training using Salient Span Masking \citep{guu2020realm}.
In all three cases, we observe improvements over the prior state-of-the-art T5-XXL model (without SSM).
Natural Questions exact match increases to 34.4 versus the prior best of 32.8, Web Questions increases to 41.0 over 37.2, and TriviaQA increases to 47.5 versus 42.9. 

Summing up, despite training on less than half the data of other models, we already find comparable, and sometimes state-of-the-art, model quality.
Currently, the Switch Transformer translates substantial upstream gains better to knowledge-based tasks, than reasoning-tasks (see Appendix 
\ref{app: upstream_downstream}).
Extracting stronger fine-tuning performance from large expert models is an active research question, and the pre-training perplexity indicates future improvements should be possible.

\section{Related Work}
The importance of scale in neural networks is widely recognized and several approaches have been proposed.
Recent works have scaled models to billions of parameters through using model parallelism (e.g. splitting weights and tensors across multiple cores) \citep{shazeer2018mesh,rajbhandari2019zero,raffel2019exploring,brown2020language,shoeybi2019megatron}.
Alternatively, \cite{harlap2018pipedream,huang2019gpipe} propose using pipeline based model parallelism, where different layers are split across devices and micro-batches are \emph{pipelined} to the different layers.
Finally, Product Key networks \citep{lample2019large} were proposed to scale up the capacity of neural networks by doing a lookup for learnable embeddings based on the incoming token representations to a given layer.

Our work studies a specific model in a class of methods that do \emph{conditional} computation, where computation decisions are made dynamically based on the input. 
\cite{cho2014exponentially} proposed adaptively selecting weights based on certain bit patterns occuring in the model hidden-states. 
\cite{eigen2013learning} built stacked expert layers with dense matrix multiplications and ReLU activations and showed promising results on jittered MNIST and monotone speech.
In computer vision \cite{puigcerver2020scalable} manually route tokens based on semantic classes during upstream pre-training and then select the relevant experts to be used according to the downstream task.

Mixture of Experts (MoE), in the context of modern deep learning architectures, was proven effective in \citet{shazeer2017outrageously}.
That work added an MoE layer which was stacked between LSTM \citep{hochreiter1997long} layers, and tokens were separately routed to combinations of experts.
This resulted in state-of-the-art results in language modeling and machine translation benchmarks.
The MoE layer was reintroduced into the Transformer architecture by the Mesh Tensorflow library \citep{shazeer2018mesh} where MoE layers were introduced as a substitute of the FFN layers, however, there were no accompanying NLP results.
More recently, through advances in machine learning infrastructure, GShard \citep{lepikhin2020gshard}, which extended the XLA compiler, used the MoE Transformer to dramatically improve machine translation across 100 languages.
Finally \cite{fan2021beyond} chooses a different deterministic MoE strategy to split the model parameters into non-overlapping groups of languages.

Sparsity along the sequence length dimension ($L$) in the Transformer \emph{attention patterns} has been a successful technique to reduce the attention complexity from $O(L^2)$ \citep{child2019generating,correia2019adaptively,sukhbaatar2019adaptive, kitaev2020reformer, zaheer2020big, beltagy2020longformer}. This has enabled learning longer sequences than previously possible. This version of the Switch Transformer does not employ attention sparsity, but these techniques are complimentary, and, as future work, these could be combined to potentially improve learning on tasks requiring long contexts.

\section{Discussion}
We pose and discuss questions about the Switch Transformer, and sparse expert models generally, where sparsity refers to weights, not on attention patterns.

\textbf{Isn't Switch Transformer better due to sheer parameter count?} Yes, and by design! Parameters, independent of the total FLOPs used, are a useful axis to scale neural language models. Large models have been exhaustively shown to perform better \citep{kaplan2020scaling}. But in this case, our model is more sample efficient and faster while using the same computational resources.

\textbf{I don't have access to a supercomputer---is this still useful for me?}
Though this work has focused on extremely large models, we also find that models with as few as two experts improves performance while easily fitting within memory constraints of commonly available GPUs or TPUs (details in Appendix \ref{app: few_experts}).
We therefore believe our techniques are useful in small-scale settings.

\textbf{Do sparse models outperform dense models on the speed-accuracy Pareto curve?}
Yes. Across a wide variety of different models sizes, sparse models outperform dense models per step and on wall clock time. Our controlled experiments show for a fixed amount of computation and time, sparse models outperform dense models.

\textbf{I can't deploy a trillion parameter model---can we shrink these models?}
We cannot fully preserve the model quality, but compression rates of 10 to 100x are achievable by distilling our sparse models into dense models while achieving $\approx$30\% of the quality gain of the expert model.

\textbf{Why use Switch Transformer instead of a model-parallel dense model?} 
On a time basis, Switch Transformers can be far more efficient than dense-models with sharded parameters (Figure~\ref{fig:dense_vs_expert_scaling}).
Also, we point out that this decision is \emph{not} mutually exclusive---we can, and do, use model-parallelism in Switch Transformers, increasing the FLOPs per token, but incurring the slowdown of conventional model-parallelism.

\textbf{Why aren't sparse models widely used already?}
The motivation to try sparse models has been stymied by the massive success of scaling dense models (the success of which is partially driven by co-adaptation with deep learning hardware as argued in \cite{hooker2020hardware}). Further, sparse models have been subject to multiple issues including (1) model complexity, (2) training difficulties, and (3) communication costs.
Switch Transformer makes strides to alleviate these issues.

\section{Future Work}
This paper lays out a simplified architecture, improved training procedures, and a study of how sparse models scale.
However, there remain many open future directions which we briefly describe here:

\begin{enumerate}
\item A significant challenge is further improving training stability for the largest models.
While our stability techniques were effective for our Switch-Base, Switch-Large and Switch-C models (no observed instability), they were not sufficient for Switch-XXL.
We have taken early steps towards stabilizing these models, which we think may be generally useful for large models, including using regularizers for improving stability and adapted forms of gradient clipping, but this remains unsolved.

\item Generally we find that improved pre-training quality leads to better downstream results (Appendix \ref{app: upstream_downstream}), though we sometimes encounter striking anomalies.
For instance, despite similar perplexities modeling the C4 data set, the 1.6T parameter Switch-C achieves only an 87.7 exact match score in SQuAD, which compares unfavorably to 89.6 for the smaller Switch-XXL model.
One notable difference is that the Switch-XXL model applies $\approx$10x the FLOPS per token than the Switch-C model, even though it has $\approx$4x less unique parameters (395B vs 1.6T).
This suggests a poorly understood dependence between fine-tuning quality, \emph{FLOPS per token} and \emph{number of parameters}.

\item Perform a comprehensive study of scaling relationships to guide the design of architectures blending data, model and expert-parallelism.
Ideally, given the specs of a hardware configuration (computation, memory, communication) one could more rapidly design an optimal model.
And, vice versa, this may also help in the design of future hardware.

\item Our work falls within the family of adaptive computation algorithms.
Our approach always used identical, homogeneous experts, but future designs (facilitated by more flexible infrastructure) could support \emph{heterogeneous} experts.
This would enable more flexible adaptation by routing to larger experts when more computation is desired---perhaps for harder examples.

\item Investigating expert layers outside the FFN layer of the Transformer.
We find preliminary evidence that this similarly can improve model quality.
In Appendix \ref{app: expert_attention}, we report quality improvement adding these inside Self-Attention layers, where our layer replaces the weight matrices which produce Q, K, V.
However, due to training instabilities with the bfloat16 format, we instead leave this as an area for future work.

\item Examining Switch Transformer in new and across different modalities. We have thus far only considered language, but we believe that model sparsity can similarly provide advantages in new modalities, as well as multi-modal networks.
\end{enumerate}

This list could easily be extended, but we hope this gives a flavor for the types of challenges that we are thinking about and what we suspect are promising future directions.

\section{Conclusion}
Switch Transformers are scalable and effective natural language learners.
We simplify Mixture of Experts to produce an architecture that is easy to understand, stable to train and vastly more sample efficient than equivalently-sized dense models.
We find that these models excel across a diverse set of natural language tasks and in different training regimes, including pre-training, fine-tuning and multi-task training.
These advances make it possible to train models with hundreds of billion to trillion parameters and which achieve substantial speedups relative to dense T5 baselines.
We hope our work motivates sparse models as an effective architecture and that this encourages researchers and practitioners to consider these flexible models in natural language tasks, and beyond.

\acks{The authors would like to thank Margaret Li who provided months of key insights into algorithmic improvements and suggestions for empirical studies.
Hugo Larochelle for sage advising and clarifying comments on the draft,
Irwan Bello for detailed comments and careful revisions,
Colin Raffel and Adam Roberts for timely advice on neural language models and the T5 code-base,
Yoshua Bengio for advising and encouragement on research in adaptive computation,
Jascha Sohl-dickstein for interesting new directions for stabilizing new large scale models and paper revisions,  and the Google Brain Team for useful discussions on the paper.
Blake Hechtman who provided invaluable help in profiling and improving the training performance of our models.}

\appendix

\section{Switch for Attention} \label{app: expert_attention}
\begin{figure}[ht!]
    \centering
    \includegraphics[width=0.85\columnwidth]{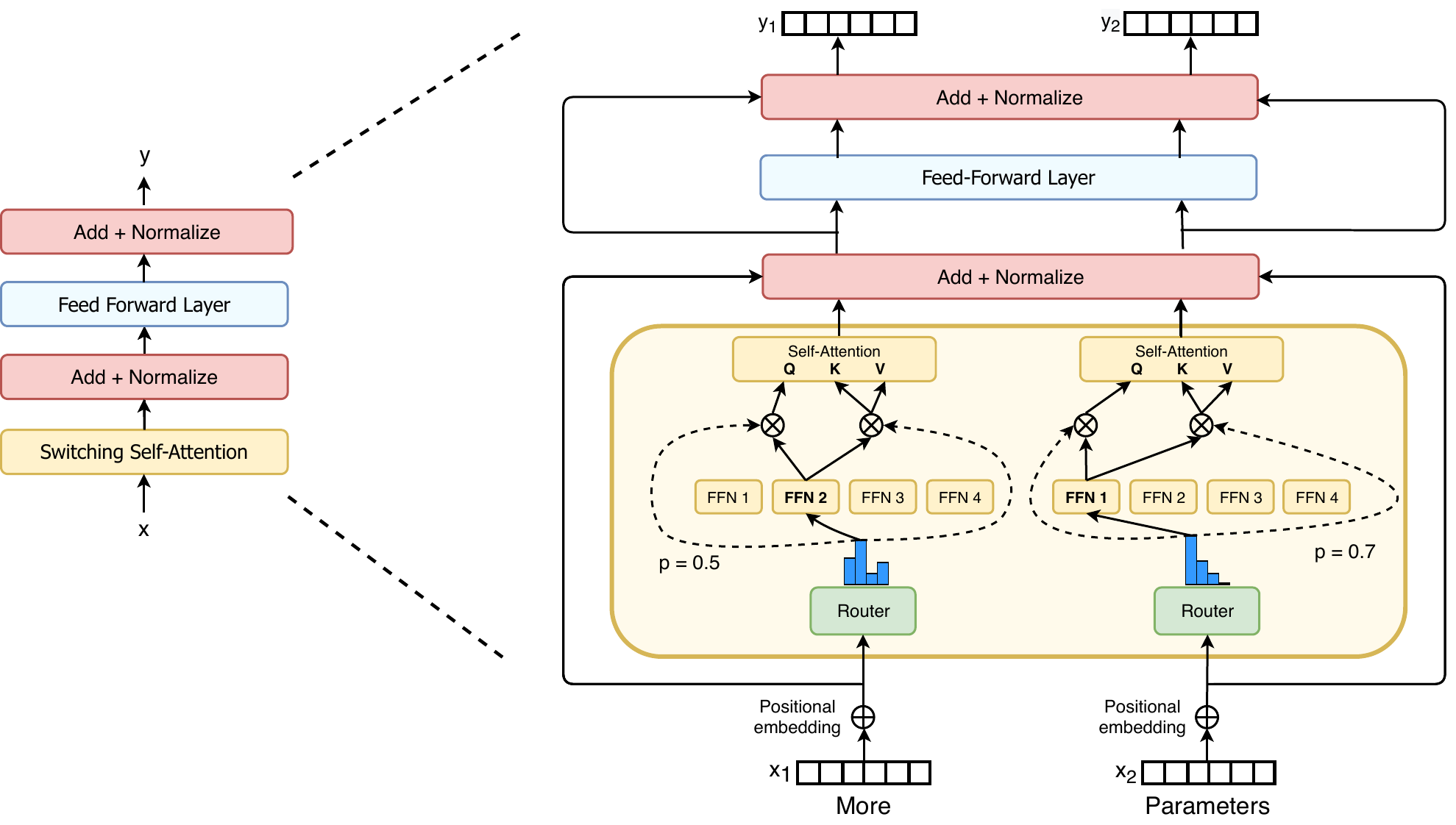}
    \caption{Switch layers in attention. We diagram how to incorporate the Switch layer into the Self-Attention transformer block. For each token (here we show two tokens, $x_1=\text{``More"}$ and $x_2=\text{``Parameters"}$), one set of weights produces the query and the other set of unique weights produces the shared keys and values. We experimented with each expert being a linear operation, as well as a FFN, as was the case throughout this work. While we found quality improvements using this, we found this to be more unstable when used with low precision number formats, and thus leave it for future work.}
    \label{fig:experts_attention}
\end{figure}

\citet{shazeer2018mesh,lepikhin2020gshard} designed MoE Transformers \citep{shazeer2017outrageously} by adding MoE layers into the dense feedfoward network (FFN) computations of the Transformer.
Similarly, our work also replaced the FFN layer in the Transformer, but we briefly explore here an alternate design.
We add Switch layers into the Transformer \emph{Self-Attention} layers.
To do so, we replace the trainable weight matrices that produce the queries, keys and values with Switch layers as seen in Figure \ref{fig:experts_attention}.

Table \ref{tab: experts_attention} records the quality after a fixed number of steps as well as training time for several variants.
Though we find improvements, we also found these layers to be more unstable when using bfloat16 precision and thus we did not include them in the final variant.
However, when these layers do train stably, we believe the preliminary positive results suggests a future promising direction.

\begin{table}[ht!]
\centering
\begin{tabular}{c|cccc}
    \toprule
    Model  & Precision & Quality & Quality & Speed \\
      &  & @100k Steps ($\uparrow$) & @16H ($\uparrow$) & (ex/sec) ($\uparrow$) \\
    \hline
    Experts FF & float32 & -1.548 & -1.614 & 1480 \\
    Expert Attention & float32 & -1.524 & \textbf{-1.606} & 1330 \\
    Expert Attention & bfloat16 & [diverges] & [diverges] & -- \\
    Experts FF + Attention & float32 & \textbf{-1.513} & -1.607 & 1240 \\
    Expert FF + Attention & bfloat16 & [diverges] & [diverges] & -- \\
    \bottomrule
\end{tabular}
\caption{Switch attention layer results. All models have 32 experts and train with 524k tokens per batch. Experts FF is when experts replace the FFN in the Transformer, which is our standard setup throughout the paper. Experts FF + Attention is when experts are used to replace both the FFN and the Self-Attention layers. When training with bfloat16 precision the models that have experts attention diverge.}
\label{tab: experts_attention}
\end{table}

\section{Preventing Token Dropping with \emph{No-Token-Left-Behind}}
Due to software constraints on TPU accelerators, the shapes of our Tensors must be statically sized. As a result, each expert has a finite and fixed capacity to process token representations.
This, however, presents an issue for our model which dynamically routes tokens at run-time that may result in an uneven distribution over experts.
If the number of tokens sent to an expert is less than the expert capacity, then the computation may simply be padded -- an inefficient use of the hardware, but mathematically correct.
However, when the number of tokens sent to an expert is larger than its capacity (expert overflow), a protocol is needed to handle this.
\citet{lepikhin2020gshard} adapts a Mixture-of-Expert model and addresses expert overflow by passing its representation to the next layer without processing through a residual connection which we also follow. 

\begin{figure}[h!]
    \centering
    \includegraphics[width=0.5\columnwidth]{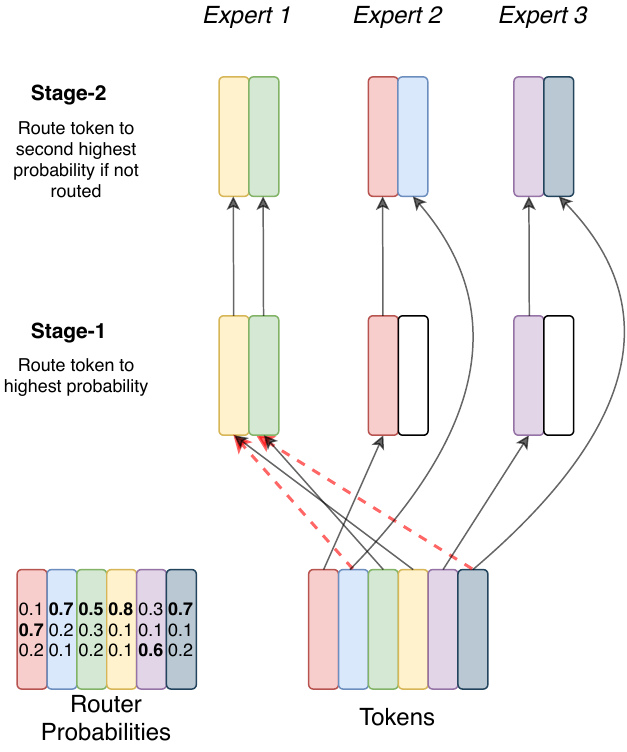}
    \caption{Diagram of the \emph{No-Token-Left-Behind Routing}. Stage 1 is equivalent to Switch routing where tokens are routed to the expert with the highest probability from the router. In Stage 2 we look at all tokens that have overflowed and route them to the expert with which has the second highest probability. Tokens can still be overflowed if their second highest expert has too many tokens, but this allows most of the tokens to be routed. This process can be iterated to guarantee virtually no tokens are dropped at all.}
    \label{fig: backup_routing}
\end{figure}

 We suspected that having no computation applied to tokens could be very wasteful, especially since if there is overflow on one expert, that means another expert will have extra capacity. With this intuition we create \emph{No-Token-Left-Behind}, which iteratively reroutes any tokens that are at first routed to an expert that is overflowing. Figure ~\ref{fig: backup_routing} shows a graphical description of this method, which will allow us to guarantee almost no tokens will be dropped during training and inference.  We hypothesised that this could improve performance and further stabilize training, but we found no empirical benefits. We suspect that once the network learns associations between different tokens and experts, if this association is changed (e.g. sending a token to its second highest expert) then performance could be degraded.

\section{Encouraging Exploration Across Experts}
At each expert-layer, the router determines to which expert to send the token.
This is a discrete decision over the available experts, conditioned on information about the token's representation.
Based on the incoming token representation, the router determines the best expert, however, it receives no counterfactual information about how well it would have done selecting an alternate expert.
As in reinforcement learning, a classic exploration-exploitation dilemma arises \citep{sutton2018reinforcement}.
These issues have been similarly noted and addressed differently by \cite{rosenbaum2017routing} which demonstrated success in multi-task learning.
This particular setting most closely matches that of a contextual bandit \citep{robbins1952some}.
Deterministically selecting the top expert always amounts to an exploitative strategy -- we consider balancing exploration to seek better expert assignment.

\begin{center}
\begin{table}[ht!]
\centering
\begin{tabular}{cc}
    \toprule
    {Model} & {Quality (Neg. Log Perp.) ($\uparrow$)}\\ \hline
    Argmax  & -1.471 \\ 
    Sample softmax  & -1.570 \\ 
    Input dropout  & -1.480 \\ 
    Input jitter  & \textbf{-1.468} \\ 
    \bottomrule
\end{tabular}
\caption{Router Exploration Strategies. Quality of the Switch Transformer, measured by the negative log perplexity, under different randomness-strategies for selecting the expert (lower is better). There is no material speed performance difference between the variants.}
\label{tab: top1_noise}
\end{table}
\end{center}

To introduce exploration, we consider several approaches: 1) deterministic or argmax 2) sampling from the softmax distribution 3) input dropout on the incoming representation 4) multiplicative jitter noise on the incoming representation.
The resulting impact on model quality is reported in Table \ref{tab: top1_noise}.
Throughout this work, we use input jitter to inject noise as we have found it to empirically perform the best.

\section{Switch Transformers in Lower Compute Regimes}\label{app: few_experts}
\begin{figure}[h!]
    \centering
    \includegraphics[width=0.65\columnwidth]{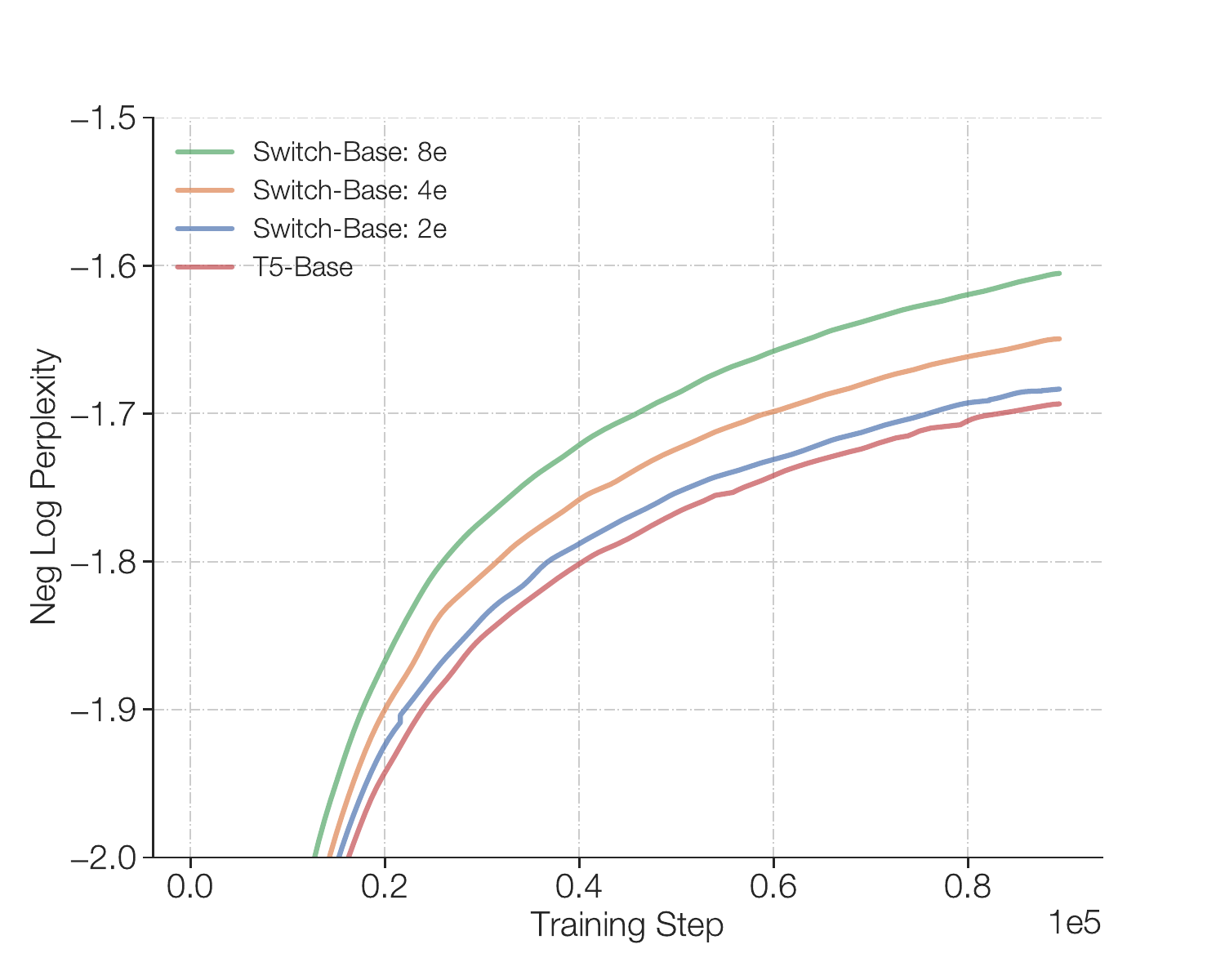}
    \caption{Switch Transformer with few experts. Switch Transformer improves over the baseline even with very few experts. Here we show scaling properties at very small scales, where we improve over the T5-Base model using 2, 4, and 8 experts.}
    \label{fig: few_experts}
\end{figure}

Switch Transformer is also an effective architecture at small scales as well as in regimes with thousands of cores and trillions of parameters.
Many of our prior experiments were at the scale of 10B+ parameter models, but we show in Figure \ref{fig: few_experts} as few as 2 experts produce compelling gains over a FLOP-matched counterpart. Even if a super computer is not readily available, training Switch Transformers with 2, 4, or 8 experts (as we typically recommend one expert per core) results in solid improvements over T5 dense baselines.

\clearpage
\section{Relation of Upstream to Downstream Model Performance} \label{app: upstream_downstream}
\begin{figure}[!ht]
    \centering
    \includegraphics[width=0.49\columnwidth]{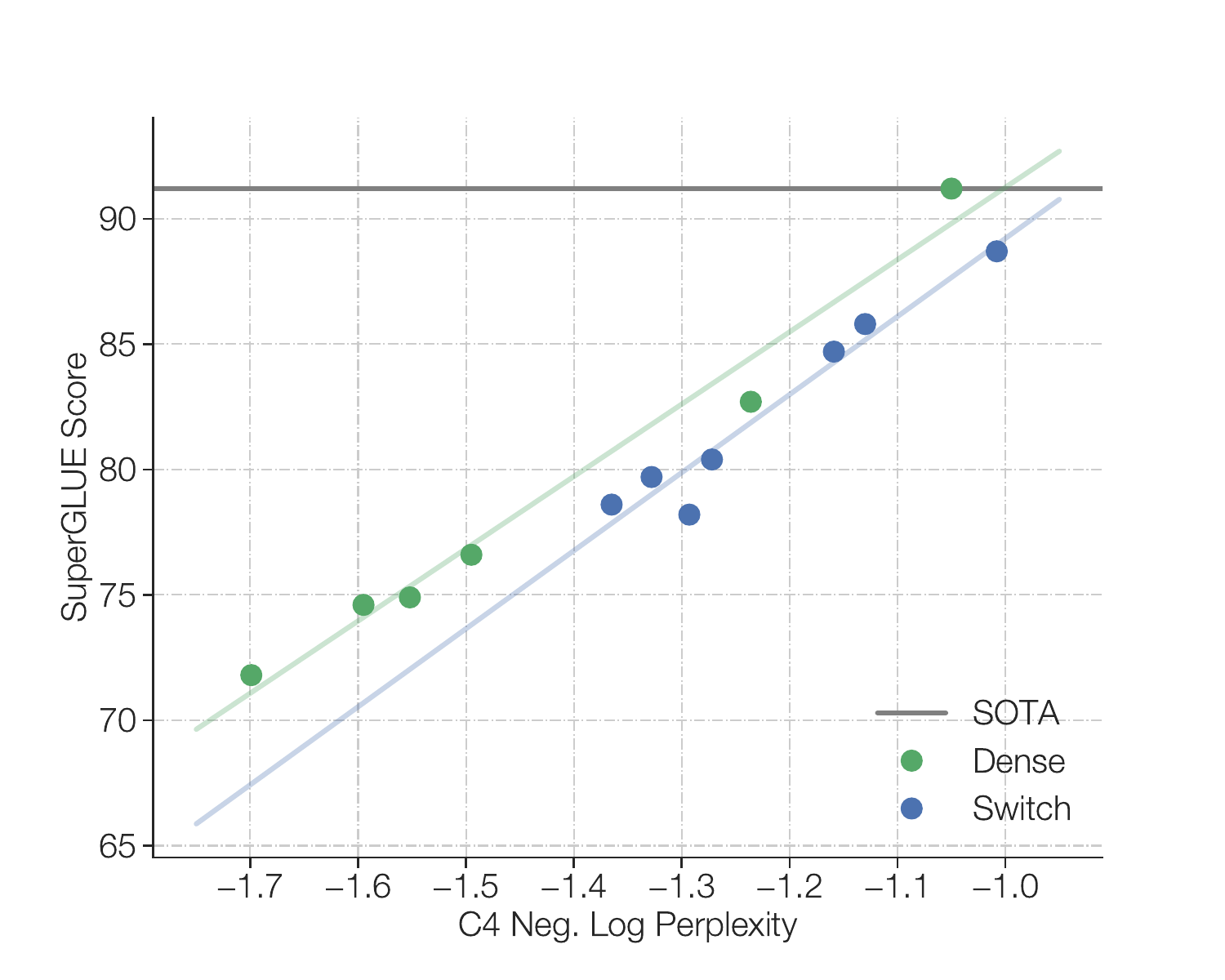}
    \includegraphics[width=0.49\columnwidth]{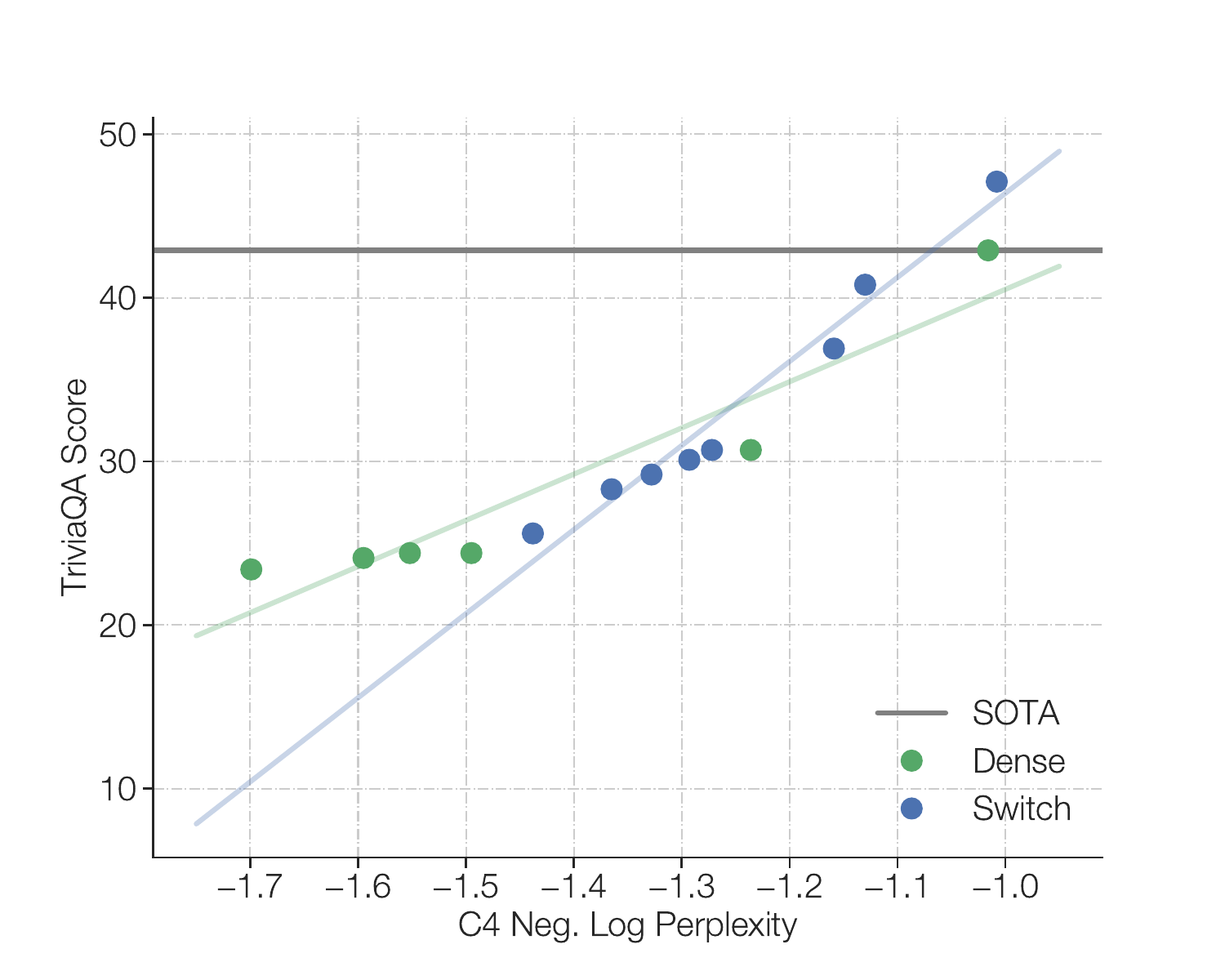}
    \caption{Upstream pre-trained quality to downstream model quality. We correlate the upstream performance with downstream quality on both SuperGLUE and TriviaQA (SOTA recorded without SSM), reasoning and knowledge-heavy benchmarks, respectively (validation sets). We find that, as with the baseline, the Switch model scales with improvements in the upstream pre-training task. For SuperGLUE, we find a loosely linear relation between negative log perplexity and the average SuperGLUE score. However, the dense model often performs better for a fixed perplexity, particularly in the large-scale regime. Conversely, on the knowledge-heavy task, TriviaQA, we find that the Switch Transformer may follow an improved scaling relationship -- for a given upstream perplexity, it does better than a dense counterpart. Further statistics (expensive to collect and left to future work) would be necessary to confirm these observations.}
    \label{fig:downstream_scaling}
\end{figure}

There is no guarantee that a model's quality on a pre-training objective will translate to downstream task results.
Figure \ref{fig:downstream_scaling} presents the correlation of the upstream model quality, for both dense and Switch models, on the C4 pre-training task with two downstream task measures:  average SuperGLUE performance and TriviaQA score.
We choose these two tasks as one probes the model's reasoning and the other factual knowledge.

We find a consistent correlation, indicating that for both baseline and Switch models, improved pre-training leads to better downstream results.
Additionally, for a fixed upstream perplexity we find that both Switch and dense models perform similarly in the small to medium model size regime.
However, in the largest model regime (T5-11B/T5-XXL) our largest Switch models, as mentioned in Section~\ref{sec:trillion_results}, do not always translate their upstream perplexity well to downstream fine-tuning on the SuperGLUE task.
This warrants future investigation and study to fully realize the potential of sparse models.
Understanding the fine-tuning dynamics with expert-models is very complicated and is dependent on regularization, load-balancing, and fine-tuning hyper-parameters.

\clearpage
\section{Pseudo Code for Switch Transformers} \label{sec:pseudo_code}
Pseudocode for Switch Transformers in Mesh Tensorflow \citep{shazeer2018mesh}. No model parallelism is being used for the below code (see~\ref{sec: expert_data_parallelism} for more details).

\begin{figure}[ht!]
\begin{lstlisting}[language=Python]
import mesh_tensorflow as mtf

def load_balance_loss(router_probs, expert_mask):
    """Calculate load-balancing loss to ensure diverse expert routing."""
    # router_probs is the probability assigned for each expert per token.
    # router_probs shape: [num_cores, tokens_per_core, num_experts]
    # expert_index contains the expert with the highest router probability in one-hot format.
    # expert_mask shape: [num_cores, tokens_per_core, num_experts]

    # For each core, get the fraction of tokens routed to each expert.
    # density_1 shape: [num_cores, num_experts]
    density_1 = mtf.reduce_mean(expert_mask, reduced_dim=tokens_per_core)
    
    # For each core, get fraction of probability mass assigned to each expert
    # from the router across all tokens.
    # density_1_proxy shape: [num_cores, num_experts]
    density_1_proxy = mtf.reduce_mean(router_probs, reduced_dim=tokens_per_core)
    
    # density_l for a single core: vector of length num_experts that sums to 1.
    # density_l_proxy for a single core: vector of length num_experts that sums to 1.
    # Want both vectors to have uniform allocation (1/num_experts) across all num_expert elements.
    # The two vectors will be pushed towards uniform allocation when the dot product is minimized.
    loss = mtf.reduce_mean(density_1_proxy * density_1) * (num_experts ^ 2)
    return loss
\end{lstlisting}
\caption{Pseudo code for the load balance loss for Switch Transformers in Mesh Tensorflow.}
\end{figure}

\begin{figure}[ht!]
\begin{lstlisting}[language=Python]
import mesh_tensorflow as mtf

def router(inputs, capacity_factor):
    """Produce the combine and dispatch tensors used for sending and 
    receiving tokens from their highest probability expert. """
    # Core layout is split across num_cores for all tensors and operations.
    # inputs shape: [num_cores, tokens_per_core, d_model]
    
    router_weights = mtf.Variable(shape=[d_model, num_experts])
    
    # router_logits shape: [num_cores, tokens_per_core, num_experts]
    router_logits = mtf.einsum([inputs, router_weights], reduced_dim=d_model)
    
    if is_training:
        # Add noise for exploration across experts.
        router_logits += mtf.random_uniform(shape=router_logits.shape, minval=1-eps, maxval=1+eps)
    
    # Convert input to softmax operation from bfloat16 to float32 for stability.
    router_logits = mtf.to_float32(router_logits)
    
    # Probabilities for each token of what expert it should be sent to.
    router_probs = mtf.softmax(router_logits, axis=-1)
    
    # Get the top-1 expert for each token. expert_gate is the top-1 probability
    # from the router for each token. expert_index is what expert each token
    # is going to be routed to.
    # expert_gate shape: [num_cores, tokens_per_core]
    # expert_index shape: [num_cores, tokens_per_core]
    expert_gate, expert_index = mtf.top_1(router_probs, reduced_dim=num_experts)
    
    # expert_mask shape: [num_cores, tokens_per_core, num_experts]
    expert_mask = mtf.one_hot(expert_index, dimension=num_experts)
    
    # Compute load balancing loss.
    aux_loss = load_balance_loss(router_probs, expert_mask)

    # Experts have a fixed capacity, ensure we do not exceed it. Construct
    # the batch indices, to each expert, with position_in_expert
    # make sure that not more that expert_capacity examples can be routed to
    # each expert.
    position_in_expert = mtf.cumsum(expert_mask, dimension=tokens_per_core) * expert_mask
    
    # Keep only tokens that fit within expert_capacity.
    expert_mask *= mtf.less(position_in_expert, expert_capacity)
    expert_mask_flat = mtf.reduce_sum(expert_mask, reduced_dim=experts_dim)

    # Mask out the experts that have overflowed the expert capacity. 
    expert_gate *= expert_mask_flat
    
    # combine_tensor used for combining expert outputs and scaling with router probability.
    # combine_tensor shape: [num_cores, tokens_per_core, num_experts, expert_capacity]
    combine_tensor = (
        expert_gate * expert_mask_flat *
        mtf.one_hot(expert_index, dimension=num_experts) *
        mtf.one_hot(position_in_expert, dimension=expert_capacity))
    
    # Cast back outputs to bfloat16 for the rest of the layer.
    combine_tensor = mtf.to_bfloat16(combine_tensor)
    
    # Create binary dispatch tensor that is 1 if the token gets routed to the corresponding expert.
    # dispatch_tensor shape: [num_cores, tokens_per_core, num_experts, expert_capacity]
    dispatch_tensor = mtf.cast(combine_tensor, tf.bool)
    
    return dispatch_tensor, combine_tensor, aux_loss
    
\end{lstlisting}
\caption{Pseudo code for the router for Switch Transformers in Mesh Tensorflow.}
\label{code: router}
\end{figure}

\begin{figure}[ht!]

\begin{lstlisting}[language=Python]
import mesh_tensorflow as mtf

def switch_layer(inputs, n, capacity_factor, num_experts):
    """Distributed switch transformer feed-forward layer."""
    # num_cores (n) = total cores for training the model (scalar).
    # d_model = model hidden size (scalar).
    # num_experts = total number of experts.
    # capacity_factor = extra buffer for each expert.
    # inputs shape: [batch, seq_len, d_model]
    batch, seq_len, d_model = inputs.get_shape()
    
    # Each core will route tokens_per_core tokens to the correct experts.
    tokens_per_core = batch * seq_len / num_cores

    # Each expert will have shape [num_cores, expert_capacity, d_model].
    # Each core is responsible for sending expert_capacity tokens
    # to each expert.
    expert_capacity = tokens_per_core * capacity_factor / num_experts

    # Reshape to setup per core expert dispatching.
    # shape: [batch, seq_len, d_model] -> [num_cores, tokens_per_core, d_model]
    # Core layout: [n, 1, 1] -> [n, 1, 1]
    inputs = mtf.reshape(inputs, [num_cores, tokens_per_core, d_model])
    
    # Core Layout: [n, 1, 1] -> [n, 1, 1, 1], [n, 1, 1, 1]
    # dispatch_tensor (boolean) shape: [num_cores, tokens_per_core, num_experts, expert_capacity]
    # dispatch_tensor is used for routing tokens to the correct expert.
    # combine_tensor (float) shape: [num_cores, tokens_per_core, num_experts, expert_capacity]
    # combine_tensor used for combining expert outputs and scaling with router
    # probability.
    dispatch_tensor, combine_tensor, aux_loss = router(inputs, expert_capacity)
    
    # Matmul with large boolean tensor to assign tokens to the correct expert. 
    # Core Layout: [n, 1, 1], -> [1, n, 1, 1]
    # expert_inputs shape: [num_experts, num_cores, expert_capacity, d_model]
    expert_inputs = mtf.einsum([inputs, dispatch_tensor], reduce_dims=[tokens_per_core])
    
    # All-to-All communication. Cores split across num_cores and now we want to split 
    # across num_experts. This sends tokens, routed locally, to the correct expert now 
    # split across different cores.
    # Core layout: [1, n, 1, 1] -> [n, 1, 1, 1]
    expert_inputs = mtf.reshape(expert_inputs, [num_experts, num_cores, expert_capacity, d_model])
    
    # Standard feed forward computation, where each expert will have its own 
    # unique set of parameters.
    # Total unique parameters created: num_experts * (d_model * d_ff * 2). 
    # expert_outputs shape: [num_experts, num_cores, expert_capacity, d_model]
    expert_outputs = feed_forward(expert_inputs)
    
    # All-to-All communication. Cores are currently split across the experts 
    # dimension, which needs to be switched back to being split across num_cores.
    # Core Layout: [n, 1, 1, 1] -> [1, n, 1, 1]
    expert_outputs = mtf.reshape(expert_outputs, [num_experts, num_cores, expert_capacity, d_model])
    
    # Convert back to input shape and multiply outputs of experts by the routing probability.
    # expert_outputs shape: [num_experts, num_cores, tokens_per_core, d_model]
    # expert_outputs_combined shape: [num_cores, tokens_per_core, d_model]
    # Core Layout: [1, n, 1, 1] -> [n, 1, 1]
    expert_outputs_combined = mtf.einsum([expert_outputs, combine_tensor], reduce_dims=[tokens_per_core])
    
    # Remove tokens_per_core shapes used for local routing dispatching to match input shape.
    # Core Layout: [n, 1, 1] -> [n, 1, 1]
    outputs = mtf.reshape(expert_outputs_combined, [batch, seq_len, d_model])
    return outputs, aux_loss


\end{lstlisting}
\caption{Pseudo code of the Switch Transformer layer in Mesh Tensorflow.}
\end{figure}

\clearpage
\bibliography{references}

\end{document}